\documentclass{article}

\usepackage[latin1]{inputenc}
\usepackage{amsfonts}
\usepackage{amssymb}
\usepackage{amsmath}
\usepackage{amsthm}
\usepackage{latexsym}
\usepackage{verbatim}
\usepackage{epsfig}
\usepackage{color}
\usepackage{url}
										
\usepackage{graphicx}
\usepackage{amsbsy}
\usepackage{amscd}
\usepackage{bm}

\newcommand{\R}{{\mathbb R}}

\DeclareMathOperator*{\tr}{tr}
\DeclareMathOperator*{\diag}{diag}
\DeclareMathOperator*{\E}{\mathbf{E}}
\DeclareMathOperator*{\argmin}{argmin}

\DeclareMathOperator*{\p}{\mathbf{P}}
\providecommand{\wt}[1]{\widetilde{#1}}
\providecommand{\wh}[1]{\widehat{#1}}

\providecommand{\norm}[1]{\left \lVert#1 \right \rVert}
\providecommand{\nnorm}[1]{ \lVert#1 \rVert}
\newcommand{\scp}[2]{\left\langle#1, #2\right\rangle}

\newcommand{\dist}[2]{\text{dist}(#1,#2)}
\providecommand{\mc}[1]{\mathcal#1}
\providecommand{\T}{\top}

\newcommand{\blanco}[1]{  }

\newcommand{\deriv}[3]{%
\ifthenelse{#1 = 1}{\frac{d\,#2}{d\,#3}}{\frac{d^{{#1}} #2}{d{#3}^{{#1}}}}
}

\newcommand{\partials}[3]{%
\ifthenelse{#1 = 1}{\frac{\partial\,#2}{\partial\,#3}}{\frac{\partial^{#1}
    #2}{\partial#3^{#1}}}
}

\def\su{\sum_{i=1}^n}

\def \coloneq{\mathrel{\mathop:}=}
\def \invcoloneq{=\mathrel{\mathop:}}
\def \eps{\varepsilon}

\def \gec{\succeq}
\def \lec{\preceq}

\newtheorem{theo}{Theorem}
\newtheorem{propo}{Theorem}

\newtheorem{conditio}{Theorem}

\newtheorem{corollary}{Theorem}

\newtheorem{definitioA}{Theorem}[section]

\newtheorem{lemmaA}{Theorem}[section]
\newtheorem{theoA}{Theorem}[section]

  \newtheorem{prop}[propo]{Proposition}
  \newtheorem{corro}[corollary]{Corollary}

\newtheorem{cond}[conditio]{Condition}

\newtheorem{defnApp}[definitioA]{Definition}
\newtheorem{lemmaApp}[lemmaA]{Lemma}
\newtheorem{theoremApp}[theoA]{Theorem}

\newenvironment{bew}{\begin{proof}[Proof]}{\end{proof}}

\usepackage{floatrow}
\usepackage{subfigure}
\newfloatcommand{capbtabbox}{table}[][\FBwidth]
\def\pa{\parallel}
\def\pe{\perp}
\def\psd{\mathbb{S}_+}
\def\sym{\mathbb{S}}
\def\TT{\mathbb{T}}

\begin{document}

\title{\begin{tabular}{l} Regularization-free estimation
    in trace regression \\ with symmetric positive semidefinite matrices
\end{tabular}
}
\author{
{\large Martin Slawski$^*$, $\;$Ping Li$^*$,
 Matthias Hein$^{\S}$}\\[1ex]
\hspace{0.03\textwidth}\begin{tabular}{lll}
$_*$ & & $_{\S}$ \\
{\normalsize Department of Statistics and Biostatistics} & $\;\;$&{\normalsize Department
of Mathematics} \\
{\normalsize Department
of Computer Science} & $\;\;$ & {\normalsize Department
of Computer Science} \\
{\normalsize Rutgers University} & $\;\;$ &{\normalsize Saarland University} \\
{\normalsize Piscataway, NJ 08854, USA} & $\;\;$ &{\normalsize Saarbr\"ucken, Germany}\\
\end{tabular}
}
\date{}
\maketitle

\begin{abstract} Over the past few years, trace regression models have received
considerable attention in the context of matrix completion, quantum state tomography,
and compressed sensing. Estimation of the underlying matrix from
regularization-based approaches promoting low-rankedness, notably nuclear norm
regularization, have enjoyed great popularity. In the present paper, we argue
that such regularization may no longer be necessary if the underlying matrix
is symmetric positive semidefinite (\textsf{spd}) and the design satisfies certain conditions. In this
situation, simple least squares estimation subject to an \textsf{spd} constraint may perform
as well as regularization-based approaches with a proper choice of the regularization parameter,
which entails knowledge of the noise level and/or tuning. By contrast, constrained least squares
estimation comes without any tuning parameter and may hence be preferred due to its simplicity.
\end{abstract}

\section{Introduction}\label{sec:introduction}

Trace regression models of the form
\begin{equation}\label{eq:tracereg}
y_i = \tr(X_i^{\T} \Sigma^*) + \eps_i, \; \, \, i=1,\ldots,n,
\end{equation}
where $\Sigma^* \in \R^{m_1 \times m_2}$ is the parameter of interest
to be estimated given measurement matrices $X_i \in \R^{m_1 \times m_2}$
and observations $y_i$ contaminated by errors $\eps_i$,
$i=1,\ldots,n$, have attracted considerable interest in high-dimensional statistical
inference, machine learning and signal processing over the past few years. Research
in these areas has focused on a setting with few measurements $n \ll m_1 \cdot m_2$ and
$\Sigma^*$ being (at least approximately) of low rank $r \ll \min\{m_1,m_2\}$. Such setting
is relevant, among others, to problems such as matrix completion \cite{CandesRecht2009, Srebro2005}, compressed
sensing \cite{CandesPlan2011, Negahban2011}, quantum state tomography \cite{Gross2010} and phase retrieval
\cite{Candes2012}. A common thread in these works is the use of the nuclear norm of a matrix
as a convex surrogate for its rank \cite{Recht2010} in regularized estimation amenable to
modern optimization techniques. This approach can be seen as natural generalization of $\ell_1$-norm
(aka lasso) regularization for the standard linear regression model \cite{Tib1996} that arises as a special case
of model \eqref{eq:tracereg} in which both $\Sigma^*$ and the measurement matrices $\{ X_i \}_{i = 1}^n$ are diagonal.
It is inarguable that in general regularization is essential if $n < m_1 \cdot m_2$. However, the situation
is less clear if $\Sigma^*$ is known to satisfy additional constraints that can be incorporated in estimation. Specifically, in
the present paper we consider the case in which $m_1 = m_2 = m$ and $\Sigma^*$ is known to be symmetric positive semidefinite (\textsf{spd}), written as $\Sigma^* \in \psd^m$ with $\psd^m$ denoting the positive semidefinite cone in the space of symmetric real-valued
$m \times m$ matrices $\sym^m$. The set $\psd^m$ deserves specific interest as it includes covariance matrices and Gram matrices in kernel-based learning methods \cite{ScholkopfSmola2002}. It is rather common for these matrices to be of low rank (at least approximately), given the widespread use of principal components analysis and low-rank kernel approximations \cite{Seeger2001}. In the present paper, we focus on the usefulness of the \textsf{spd} constraint for estimation. We argue that if $\Sigma^*$ is \textsf{spd} and the measurement matrices $\{ X_i \}_{i = 1}^n$ obey certain conditions, constrained least squares estimation
\begin{equation}\label{eq:constrainedleastsquares}
\min_{\Sigma \in \psd^m} \; \frac{1}{2n} \sum_{i = 1}^n (y_i - \tr(X_i^{\T} \Sigma))^2
\end{equation}
may perform similarly well in prediction and parameter estimation as approaches
employing nuclear norm regularization with proper choice of the regularization parameter, including the
interesting regime $n < \delta_m$, where $\delta_m = \dim(\sym^m) = m(m+1)/2$. Note that the objective in \eqref{eq:constrainedleastsquares} only consists of a data
fitting term and is hence convenient to work with in practice since one does not need to choose any parameter. Our findings
can be seen as a non-commutative extension of recent results on non-negative least squares estimation for high-dimensional linear regression
with non-negative parameters \cite{Meinshausen2013, SlawskiHein2013a}. In these papers it is shown that for certain design matrices, non-negative least squares can achieve comparable performance to $\ell_1$-norm regularized estimation with regard to prediction, estimation and support recovery, thereby generalizing prior work \cite{Bruckstein2008, DonohoTanner2010, WangTang2009} on sparse recovery of a non-negative
vector in a noiseless setting.

\textbf{Related work.} Model \eqref{eq:tracereg} with $\Sigma^* \in \psd^m$ has been studied in
several recent papers. A good deal of these papers consider the setup of compressed sensing according
to which the matrices $\{ X_i \}_{i = 1}^n$ can be chosen by the user, with the goal to minimize the
number of observations required to (approximately) recover $\Sigma^*$.

In \cite{Wang2011}, the problem of exactly recovering $\Sigma^*$ being low-rank
from noiseless observations ($\eps_i = 0$, $i=1,\ldots,n$) by solving a linear feasibility problem
over the positive semidefinite cone is considered, which is equivalent to the proposed
least squares problem \eqref{eq:tracereg} in a noiseless setting. Apart from the fact that
we primarily study a noisy setting, we shall argue below that in the setup of compressed sensing the measurement matrices studied in \cite{Wang2011}
constitute an unfavourable choice relative to those recommended in the present paper.

In \cite{Chen2013}, recovery from rank-one measurements is considered, i.e., ~for $\{ x_i \}_{i = 1}^n \subset \R^m$
\begin{equation}\label{eq:rankone}
y_i = x_i^{\T}  \Sigma^* x_i + \eps_i = \tr(X_i^{\T} \Sigma^*) + \eps_i, \;\;\; \text{with} \; X_i = x_i x_i^{\T}, \; i=1,\ldots,n.
\end{equation}
As opposed to \cite{Chen2013}, where estimation based on nuclear norm regularization is proposed,
the present work is devoted to regularization-free estimation. While rank-one measurements as in \eqref{eq:rankone} are
also in the center of interest herein, our framework is not limited to this specific case.

In \cite{Cai2014}, rank-one measurements are considered for general $\Sigma^* \in \R^{m_1 \times m_2}$. Specializing to
$\Sigma^* \in \psd^m$, the authors discuss an application of \eqref{eq:rankone} to covariance matrix estimation given only one-dimensional
projections $\{ x_i^{\T} z_i \}_{i = 1}^n$ of the data points, where the $\{ z_i \}_{i = 1}^n$ are i.i.d.~from a distribution with zero mean and covariance matrix
$\Sigma^*$. In fact, when using observations $y_i = (x_i^{\T} z_i)^2$, one obtains
\begin{equation}\label{eq:rankone_onedimproj}
(x_i^{\T} z_i)^2 = x_i^{\T} z_i z_i^{\T} x_i = x_i^{\T} \Sigma^* x_i + \eps_i, \; \; \text{with} \; \eps_i = x_i^{\T} \{ z_i z_i^{\T} - \Sigma^* \} x_i, \; i=1,\ldots,n.
\end{equation}
On the other hand, in \cite{Cai2014}, no specific attention is given to the \textsf{spd} constraint: the convex program proposed therein, which can be seen
as a  modification of the approach in \cite{Chen2013}, applies to general symmetric matrices and does not enforce positive semidefiniteness.

Specializing model \eqref{eq:rankone} further to the case in which also $\Sigma^* = \sigma^* (\sigma^*)^{\T}$ has rank one, one obtains the quadratic model
\begin{align}\label{eq:quadraticmodel}
y_i = |x_i^{\T} \sigma^*|^2 + \eps_i
\end{align}
which (with complex-valued $\sigma^*$) is relevant to the problem of phase retrieval \cite{Klibanov} that has received some attention recently. The approach
of \cite{Candes2012} treats \eqref{eq:quadraticmodel} as an instance of \eqref{eq:tracereg} and uses nuclear norm regularization to enforce rank-one solutions.
In follow-up work \cite{CandesLi2013}, the authors show a refined
recovery result stating that imposing an \textsf{spd} constraint $-$ without regularization $-$ suffices. A similar result
has been proven independently by \cite{Demanet2014}. However, the results in both \cite{CandesLi2013} and \cite{Demanet2014} only concern model \eqref{eq:quadraticmodel}.\\
In \cite{Koltchinskii2011}, $\Sigma^*$ is
assumed to be a complex Hermitian positive semidefinite matrix of unit trace, which is the
setting in quantum state tomography. While the setting as well as the measurement matrices
under consideration are different from ours, a notable point of contact to our work can be
seen in the fact that the negative von Neumann entropy\footnote{The von Neumann entropy of a positive definite Hermitian matrix is given by the entropy of its eigenvalues}, which is the proposed regularizer in \cite{Koltchinskii2011},
does not promote low rankedness, but constitutes one possible way of enforcing positive definiteness. At the same time,
adaptivity of the approach to low rankedness is established in \cite{Koltchinskii2011}.

\textbf{Outline and contributions of the paper.} In Section \ref{sec:analysis}, we study statistical
properties of constrained least squares estimation \eqref{eq:constrainedleastsquares} in small sample ($n < \delta_m$)
and low-rank settings. Specifically, we introduce certain geometric conditions associated with the measurements
$\{X_i\}_{i = 1}^n$ that allow us to derive non-asymptotic upper bounds on the prediction and estimation error indicating
that \eqref{eq:constrainedleastsquares} can achieve competitive performance while being regularization-free. On
the other hand, we show that without extra conditions on the measurements $\{X_i\}_{i = 1}^n$, the performance of
\eqref{eq:constrainedleastsquares} can be as poor as that of unconstrained least squares. Section \ref{sec:numerical}
contains numerical results based on synthetic and real world data that support or complement our theoretical results.
Our findings  are briefly summarized in Section \ref{sec:summary}. The appendix contains the proofs.

\textbf{Notation.} We here gather notation and terminology used throughout the paper.
For an integer $d \geq 1$, let $\mathbb{M}^d$ denote the Euclidean vector
space of real $d \times d$ matrices with inner product $\scp{M}{M'} \coloneq \tr(M^{\T} M')$,
$M,M' \in \mathbb{M}^d$. The set of real symmetric $d \times d$ matrices $\sym^d$ is a subspace
of  $\mathbb{M}^d$ of dimension $\delta_d \coloneq d (d + 1)/2$. Each element $M$ of $\sym^d$ has an eigen-decomposition
$M = U \Lambda U^{\T} = \sum_{j = 1}^d \lambda_j(M) u_j u_j^{\T}$, where $\lambda_1(M) = \lambda_{\max}(M) \geq \lambda_2(M) \geq \ldots \geq
\lambda_d(M) = \lambda_{\min}(M)$ is the sequence of real eigenvalues with corresponding orthonormal eigenvectors $\{ u_j \}_{j = 1}^d$,
$\Lambda = \diag(\lambda_1(M),\ldots,\lambda_d(M))$, and $U = [u_1 \, \ldots \, u_d]$. For $q \in [1,\infty]$, $\sym^d$ can be endowed
with a norm given by the mapping
$M \mapsto \nnorm{M}_q \coloneq \left(\sum_{j = 1}^d |\lambda_j(M)|^q \right)^{1/q}$ called the Schatten-$q$-norm. In particular,
for $q = 1$ we speak of the nuclear norm, while $q = 2$ yields the Frobenius norm $\nnorm{\cdot}_F$. We set $\nnorm{M}_{\infty} \coloneq \max_{1 \leq j \leq d} |\lambda_j(M)|$, the spectral norm  of $M$. We denote by $\mc{S}_1(d) = \{M \in \sym^d: \nnorm{M}_q = 1 \}$
the Schatten-$1$-norm unit sphere and set $\mc{S}_1^+(d) = \mc{S}_1(d) \cap \psd^d$, where
$\psd^d = \{M \in \sym^d: v^{\T} M v \geq 0 \; \forall v \in \R^d \}$ is the positive semidefinite cone in $\sym^d$.
The symbols $\gec, \lec, \succ, \prec$ are understood with respect to the semidefinite ordering, e.g.~$M \lec M'$ means that $M' - M \in \psd^d$. For $v \in \R^d$ and $q \in [1,\infty]$, $\nnorm{v}_q$ denotes the usual $q$-norm. For set $A, B$ and a real number $\alpha$, $\alpha A \coloneq \{\alpha a, a \in A \}$, $A - B = \{a - b, a \in A, b \in B \}$, and
for $A, B \subset \R^d$, $\dist{A}{B} \coloneq \min_{a \in A, b \in B} \nnorm{a - b}_2$.

It is convenient to re-write model \eqref{eq:tracereg} as
\begin{equation*}
y = \mc{X}(\Sigma^*) + \eps,
\end{equation*}
where $y = (y_i)_{i=1}^n$, $\eps = (\eps_i)_{i=1}^n$ and $\mc{X}:\mathbb{M}^m \rightarrow \R^n$ is a linear
map defined by $(\mc{X}(M))_i = \tr(X_i^{\T} M)$,
$i = 1,\ldots,n$, referred to as \emph{sampling operator}. Its adjoint $\mc{X}^*:\R^n \rightarrow \mathbb{M}^m$ is given by the map
$v \mapsto \sum_{i = 1}^n v_i X_i$.

\section{Analysis}\label{sec:analysis}

\textbf{Preliminaries.} Throughout this section, we consider a special instance of model \eqref{eq:tracereg} in which
\begin{equation}\label{eq:tracereg_spd}
y_i = \tr(X_i \Sigma^*) + \eps_i, \quad \; \text{where} \; \, \Sigma^* \in \psd^m, \; \, X_i \in \sym^m, \; \text{and} \; \,
\eps_i \overset{\text{i.i.d.}}{\sim} N(0, \sigma^2), \, i=1,\ldots,n.
\end{equation}
The assumption that the errors $\{ \eps_i \}_{i = 1}^n$ follow a Gaussian distribution is made for convenience as it
simplifies the stochastic part of our analysis, which could be extended to cover error distributions with sub-Gaussian tails.

Note that without loss of generality, we may assume that the $\{ X_i \}_{i = 1}^n$ are symmetric. In fact, any $M \in \mathbb{M}^m$
can be decomposed as
\begin{equation*}
M = M^{\text{sym}} + M^{\text{skew}}, \quad \text{where} \; \, M^{\text{sym}} = \frac{M + M^{\T}}{2} \; \text{and} \; \, M^{\text{skew}} = \frac{M - M^{\T}}{2}
\end{equation*}
denote the Euclidean projections of $M$ onto $\sym^m$ and its orthogonal complement (the subspace of skew-symmetric matrices), respectively. Accordingly, since $\Sigma^* \in \sym^m$, we have $\tr(M \Sigma^*) = \tr(M^{\text{sym}} \Sigma^*)$.

In the sequel, we study the statistical performance of the constrained least squares estimator
\begin{equation}\label{eq:constrainedleastsquares_compact}
\wh{\Sigma} \in \argmin_{\Sigma \in \psd^m} \, \frac{1}{2n} \nnorm{y - \mc{X}(\Sigma)}_2^2
\end{equation}
under model \eqref{eq:tracereg_spd} with respect to prediction and estimation. More specifically, under certain conditions on $\mc{X}$, we
shall derive bounds on
\begin{equation}\label{eq:errormeas}
(a) \;\; \frac{1}{n} \nnorm{\mc{X}(\Sigma^*) - \mc{X}(\wh{\Sigma})}_2^2,  \quad \; \text{and}  \quad \; (b) \; \; \nnorm{\wh{\Sigma} - \Sigma^*}_1,
\end{equation}
where $(a)$ will be referred to as ``prediction error'' below.

The most basic method for estimating $\Sigma^*$ is ordinary least squares (ols) estimation
\begin{equation}\label{eq:ols}
\wh{\Sigma}^{\text{ols}} \in \argmin_{\Sigma \in \sym^m} \, \frac{1}{2n} \nnorm{y - \mc{X}(\Sigma)}_2^2,
\end{equation}
which is computationally much simpler than \eqref{eq:constrainedleastsquares_compact}. While
obtaining \eqref{eq:constrainedleastsquares_compact} requires techniques from convex programming,
it is straightforward to compute \eqref{eq:ols} by solving a linear system of equations in $\delta_m = m(m+1)/2$
variables. On the other hand, the prediction error of ols scales as $O_{\p}(\dim(\text{range}(\mc{X}))/n)$, where
$\dim(\text{range}(\mc{X}))$ can be as large as $\min\{n, \delta_m \}$, in which case the
prediction error vanishes asymptotically only if $\delta_m / n \rightarrow 0$ as $n \rightarrow \infty$. Moreover,
the estimation error $\nnorm{\wh{\Sigma}^{\text{ols}} - \Sigma^*}_1$ is unbounded unless $n \geq \delta_m$. Research
conducted over the past few years has consequently focused on methods that deal successfully with the situation
$n < \delta_m$ if the target $\Sigma^*$ possesses additional structure, notably low-rankedness. Indeed, if
$\Sigma^*$ has rank $r \ll m$, the intrinsic dimension of the problem becomes (roughly) $m r \ll  \delta_m$. Rank-constrained
estimation or regularized estimation with the matrix rank as regularizer yield computationally intractable optimization
problems in general. In a large body of work, nuclear norm regularization, which can be seen  as a convex surrogate of rank regularization, is considered as a computationally convenient alternative for which a series of adaptivity properties
to underlying low-rankedness has been established, e.g.~\cite{CandesPlan2011, KLT2011, Negahban2011, Recht2010, Rohde2011}. Complementing \eqref{eq:ols} with
nuclear norm regularization gives rise to the estimator
\begin{equation}\label{eq:ols_nucnormreg}
\wh{\Sigma}^{1} \in \argmin_{\Sigma \in \sym^m} \, \frac{1}{2n} \nnorm{y - \mc{X}(\Sigma)}_2^2 + \lambda \nnorm{\Sigma}_1,
\end{equation}
where $\lambda > 0$ is a regularization parameter. In case an \textsf{spd} constraint is imposed \eqref{eq:ols_nucnormreg} becomes
\begin{equation}\label{eq:nnls_tracenormreg}
\wh{\Sigma}^{1+} \in \argmin_{\Sigma \in \psd^m} \, \frac{1}{2n} \nnorm{y - \mc{X}(\Sigma)}_2^2 + \lambda \tr(\Sigma).
\end{equation}
Our analysis aims at elucidating potential advantages of the \textsf{spd} constraint in the constrained least squares problem
\eqref{eq:constrainedleastsquares_compact} from a statistical point of view. It turns out that
depending on properties of $\mc{X}$, the behaviour of $\wh{\Sigma}$ can range from a performance similar
to the least squares estimator $\wh{\Sigma}^{\text{ols}}$ on the one hand to a performance similar to the
nuclear norm regularized estimator $\wh{\Sigma}^{1+}$ with properly chosen/tuned $\lambda$ on the other hand. The latter
case appears to be remarkable inasmuch as $\wh{\Sigma}$ may enjoy similar adaptivity properties as nuclear norm regularized
estimators even though $\wh{\Sigma}$ is obtained from a pure data fitting problem without any explicit form of regularization.
%
\subsection{Negative results}

We first discuss examples of $\mc{X}$ for which the \textsf{spd}-constrained estimator $\wh{\Sigma}$ does not improve
(substantially) over the unconstrained estimator $\wh{\Sigma}^{\text{ols}}$. At the same time, these examples provide some clues on
conditions that need to be imposed on $\mc{X}$ to achieve substantially better performance.

\textbf{Example 1: equivalence of constrained and unconstrained least squares}\\
Let $m$ be even and consider measurement matrices of the form
\begin{equation*}
X_i = \begin{bmatrix}
       \wt{X}_i &  0\\
         0      &  -\wt{X}_i
      \end{bmatrix}
\end{equation*}
for matrices $\wt{X}_i \in \sym^{m/2}$, $i=1,\ldots,n$. For $\Sigma \in \sym^m$ arbitrary, we
can partition
\begin{equation*}
\Sigma = \begin{bmatrix}
       \Sigma_{11} &  \Sigma_{12} \\
         \Sigma_{12}      &  \Sigma_{22},
      \end{bmatrix}
\end{equation*}
where $\Sigma_{11}$ is the top $m/2 \times m/2$ block of $\Sigma$ etc. We have
\begin{equation*}
\tr(X_i \Sigma) = \tr(\wt{X}_i (\Sigma_{11} - \Sigma_{22})), \quad i=1,\ldots,n.
\end{equation*}
Hence $\Sigma$ enters the least squares objective \eqref{eq:constrainedleastsquares} via
the difference of the top and bottom $m/2 \times m/2$ blocks. Since for any dimension $d$
\begin{equation*}
\{\Sigma - \Sigma':\, \Sigma \in \psd^{d}, \; \Sigma' \in \psd^d \} = \sym^d = \{\Sigma - \Sigma':\, \Sigma \in \sym^{d},\; \Sigma' \in \sym^d \},
\end{equation*}
the \textsf{spd} constraint becomes vacuous and can be dropped from \eqref{eq:constrainedleastsquares_compact}.

\textbf{Example 2: Orthonormal design}\\
The following statement indicates that for orthonormal design, the prediction error of $\wh{\Sigma}$ cannot be
expected to improve over that of $\wh{\Sigma}^{\text{ols}}$ by substantially more than a constant factor $1/2$.
\begin{prop}\label{prop:orthonormal}
Let $\Sigma^* = 0$ so that $y = \eps$, let $n = \delta_m$ and let $\{ X_i \}_{1 \leq i \leq \delta_m}$ be an orthonormal basis of $\sym^m$.
Then, $\nnorm{\mc{X}(\wh{\Sigma})}_2^2/n \rightarrow \frac{\sigma^2}{2}$ in probability as $m,n
\rightarrow \infty$.
\end{prop}
By contrast, it is desired that $\nnorm{\mc{X}(\wh{\Sigma})}_2^2/n = o_{\p}(1)$ as $m,n \rightarrow \infty$.

\textbf{Example 3: Random Gaussian design}\\
Consider the Gaussian orthogonal ensemble (GOE) of random matrices
\begin{align*}
\text{GOE}(m) = \{ X = (x_{jk})_{1 \leq j,k \leq m}, \; &\{ x_{jj} \}_{j = 1}^m \overset{\text{i.i.d.}}{\sim} N(0,1), \\
 &\{ x_{jk}=x_{kj} \}_{1 \leq j < k \leq m} \overset{\text{i.i.d.}}{\sim} N(0,1/2) \}.
\end{align*}
Random Gaussian measurements are common in compressed sensing-type settings, see e.g.~\cite{CandesPlan2011, Negahban2011}. It is
hence of interest to study measurements $X_i \overset{\text{i.i.d.}}{\sim} \text{GOE}(m)$, $i=1,\ldots,n$, in the context of the
constrained least squares problem \eqref{eq:constrainedleastsquares_compact}. The following statement, which follows from results in \cite{Amelunxen2014}, points to a serious limitation associated with the use of such measurements.
\begin{prop}\label{prop:Amelunxen} Consider measurements $X_i \overset{\text{i.i.d.}}{\sim} \text{GOE}(m)$, $i=1,\ldots,n$. Then, for any
$\eps > 0$, if $n \leq (1 - \eps) \delta_m/2$, with probability at least $1 - 32 \exp(-\eps^2 \delta_m)$, there exists $\Delta \in \psd^m$, $\Delta \neq 0$ such that $\mc{X}(\Delta) = 0$.
\end{prop}
Proposition \ref{prop:Amelunxen} has the following implications.
\begin{itemize}
\item If the number of measurements drops below one half of the ambient dimension $\delta_m$, estimating
      $\Sigma^*$ based on \eqref{eq:constrainedleastsquares_compact} becomes ill-posed; the estimation error
      $\nnorm{\wh{\Sigma} - \Sigma^*}_1$ is unbounded, irrespective of the rank of $\Sigma^*$.
\item Geometrically, the consequence of Proposition \ref{prop:Amelunxen} is that the convex cone
      $\mc{C}_{\mc{X}} = \{z \in \R^n:\, z = \mc{X}(\Delta), \; \Delta \in \psd^m \}$ contains $0$. Unless $0$ is
      contained in the boundary of $\mc{C}_{\mc{X}}$ (we conjecture that this event has measure zero), this means
      that $\mc{C}_{\mc{X}} = \R^n$, i.e., ~the \textsf{spd} constraint becomes vacuous.
\end{itemize}
\emph{Remarks.}
\begin{enumerate}
\item In \cite{Wang2011}, the following noiseless analog to the constrained least squares problem \eqref{eq:constrainedleastsquares_compact} is considered:
\begin{equation}\label{eq:psd_feas}
\text{find} \; \Sigma \in \psd^m \quad \text{such that} \, \mc{X}(\Sigma) = y = \mc{X}(\Sigma^*),
\end{equation}
where $X_i \sim \text{GOE}(m)$, $i=1,\ldots,n$. The authors prove that for all $\xi \in (0,1)$, there exists
$\alpha \in (0,1)$ so that if $n \geq \alpha \delta_m$, $\Sigma^*$ is the unique solution of the feasibility problem \eqref{eq:psd_feas} as long as $\text{rank}(\Sigma^*) \leq \xi m$. While this implies that the \textsf{spd} constraint allows undersampling (i.e., ~$n < \delta_m$), it is
not clear to what extent undersampling is possible, i.e., ~how small $\alpha$ could possibly be. Proposition \ref{prop:Amelunxen} yields
that $\alpha$ cannot be smaller than $1/2$.
\item It is of interest to relate Proposition \ref{prop:Amelunxen} to corresponding results on the vector case (equivalent to
      having diagonal $\{X_i \}_{i = 1}^n$ and diagonal $\Sigma^*$) in \cite{DonohoTanner2010}. Compared to Proposition \ref{prop:Amelunxen}, the corresponding result in \cite{DonohoTanner2010} applies to a much wider class of random measurement matrices including all random matrices with i.i.d.~entries from a symmetric distribution around zero. It is thus natural to ask whether Proposition \ref{prop:Amelunxen} holds more generally for all Wigner matrices \cite{Tao2012}.
\item The fact that the threshold $\frac{1}{2} \delta_m$ for the number measurements in Proposition \ref{prop:Amelunxen} equals
      (up to the scaling factor $\sigma^2$) the asymptotic prediction error of Example 2 is not a coincidence; this is part of
      a wider phenomenon as pointed out in \cite{Amelunxen2014}. In the framework of \cite{Amelunxen2014}, $\frac{1}{2} \delta_m$ is
      the ``statistical dimension'' of $\psd^m$.
\end{enumerate}
\subsection{Slow rate bound on the prediction error}

We now turn to the first positive result on the \textsf{spd}-constrained least squares estimator $\wh{\Sigma}$ under an additional
condition on the sampling operator $\mc{X}$. Specifically, the prediction error will be bounded as
\begin{equation}\label{eq:slowrate_overview}
\frac{1}{n} \nnorm{\mc{X}(\Sigma^*) - \mc{X}(\wh{\Sigma})}_2^2 = O(\lambda_0 \nnorm{\Sigma^*}_1 + \lambda_0^2), \quad \text{where} \; \;
\lambda_0 = \frac{1}{n} \nnorm{\mc{X}^*(\eps)}_{\infty}
\end{equation}
with $\lambda_0$ typically being of the order $O(\sqrt{m/n})$ (up to logarithmic factors). The rate in \eqref{eq:slowrate_overview}
can be a significant improvement of what is achieved by $\wh{\Sigma}^{\text{ols}}$ if $\nnorm{\Sigma^*}_1 = \tr(\Sigma^*)$ is small. If
$\lambda_0 = o(\nnorm{\Sigma^*}_1)$ that rate coincides with those of the nuclear norm regularized estimators \eqref{eq:ols_nucnormreg},
\eqref{eq:nnls_tracenormreg} with regularization parameter $\lambda \geq \lambda_0$, cf.~Theorem 1 in \cite{Rohde2011}. For nuclear norm
regularized estimators, the rate $O(\lambda_0 \nnorm{\Sigma^*}_1)$ is achieved for any choice of $\mc{X}$ and is hence slow in the sense
that the squared prediction error only decays at the rate $n^{-1/2}$ instead of $n^{-1}$. Therefore, we refer to \eqref{eq:slowrate_overview} as ``slow rate bound''.

\textbf{Condition on $\mc{X}$.} In order to arrive at a suitable condition to be imposed on $\mc{X}$ so that
\eqref{eq:slowrate_overview} can be achieved, it makes sense to re-consider Example 3 to identify possible obstacles. Proposition
\ref{prop:Amelunxen} states that as long as $n$ is bounded away from $\delta_m/2$ from above, there is a non-trivial $\Delta \in
\psd^m$ such that $\mc{X}(\Delta) = 0$. Equivalently,
\begin{align*}\label{eq:zeroincvxhull}
\begin{split}
&\dist{\mc{P}_{\mc{X}}}{0} = \min_{\Delta \in \mc{S}_1^+(m)}\nnorm{\mc{X}(\Delta)}_2 = 0, \; \; \text{where} \\
&\mc{P}_{\mc{X}} \coloneq \{z \in \R^n:\; z = \mc{X}(\Delta), \; \Delta \in \mc{S}_1^+(m) \}, \; \; \, \text{and} \; \, \mc{S}_1^+(m) \coloneq \{\Delta \in \psd^m:\, \tr(\Delta) = 1 \}.
\end{split}
\end{align*}
In this situation, it is in general not possible to derive a non-trivial upper bound on the prediction error as
$\dist{\mc{P}_{\mc{X}}}{0} = 0$ may imply that $\mc{C}_{\mc{X}} = \R^n$ in which case $\nnorm{\mc{X}(\Sigma^*) - \mc{X}(\wh{\Sigma})}_2^2 = \nnorm{\eps}_2^2$. To rule this out, the condition $\dist{\mc{P}_{\mc{X}}}{0} > 0$ appears to be a natural requirement. More strongly, one may ask for the following:
\begin{equation}\label{eq:selfreg}
\text{There exists a constant} \; \tau > 0 \; \text{such that} \; \tau_0^2(\mc{X}) \coloneq \min_{\Delta \in \mc{S}_1^+( m)} \frac{1}{n} \nnorm{\mc{X}(\Delta)}_2^2  \geq \tau^2.
\end{equation}
This condition is sufficient to obtain a slow rate bound in the vector case, cf.~Theorem 1 in \cite{SlawskiHein2013a}. However,
the condition required for the slow rate bound in Theorem \ref{theo:slowratebound} below is somewhat stronger than \eqref{eq:selfreg}.
\begin{cond}\label{cond:slowrate} \hfill \\ There exist constants $R_* > 1$ and $\tau_* > 0$ such that $\tau^2(\mc{X}, R_*) \geq \tau_*^2$, where for $R \in \R$
\begin{equation*}
\tau^2(\mc{X}, R) = \text{\emph{dist}}^2(R \mc{P}_{\mc{X}}, \mc{P}_{\mc{X}})/n = \min_{\substack{A \in R \mc{S}_1^+(m) \\
B \in \mc{S}_1^+(m)}} \, \frac{1}{n} \nnorm{\mc{X}(A) - \mc{X}(B)}_2^2.
\end{equation*}
\end{cond}
It follows from
\begin{align}\label{eq:selfreg_weaker}
\begin{split}
\tau^2(\mc{X}, R) &= \min_{\substack{A \in R \mc{S}_1^+(m) 
B \in \mc{S}_1^+(m)}} \, \frac{1}{n} \nnorm{\mc{X}(A) - \mc{X}(B)}_2^2  \\
&\leq \min_{A \in \mc{S}_1^+(m)} \, \frac{1}{n} \nnorm{\mc{X}(R \cdot A) - \mc{X}(A)}_2^2 \\
&= (R-1)^2 \, \min_{A \in \mc{S}_1^+(m)} \frac{1}{n} \nnorm{\mc{X}(A)}_2^2 =(R-1)^2 \tau_0^2(\mc{X})
\end{split}
\end{align}
that Condition \ref{cond:slowrate} is in fact stronger than \eqref{eq:selfreg}. Below, we provide
a sufficient condition on $\mc{X}$ that implies Condition \ref{cond:slowrate}.
\begin{prop}\label{prop:measurements_deterministic} Suppose that there exists $a \in \R^n$, $\nnorm{a}_2 \leq 1$, and
constants $0 < \phi_{\min} \leq \phi_{\max}$ such that
\begin{equation*}
\lambda_{\min}(n^{-1/2}\mc{X}^*(a)) \geq \phi_{\min}, \quad \text{and} \; \; \, \lambda_{\max}(n^{-1/2} \mc{X}^*(a) ) \leq \phi_{\max}.
\end{equation*}
Then for any $\zeta > 1$, $\mc{X}$ satisfies Condition \ref{cond:slowrate} with $R_* = \zeta \frac{\phi_{\max}}{\phi_{\min}}$
and $\tau_*^2 = (\zeta - 1)^2 \phi_{\max}^2$.
\end{prop}
The condition of Proposition \ref{prop:measurements_deterministic} can be phrased as having
a positive definite matrix in the unit ball of the range of $\mc{X}^*$, which, after scaling
by $1/\sqrt{n}$, has its smallest eigenvalues bounded away from zero and condition number bounded
from above. As a simple example, suppose that $X_1 = \sqrt{n} I$. Invoking Proposition \ref{prop:measurements_deterministic}
with $a = (1, 0, \ldots, 0)^{\T}$ and $\zeta = 2$,  we find that Condition \ref{cond:slowrate} is
satisfied with $R_{*} = 2$ and $\tau_*^2 = 1$. A more interesting example is random design where the
$\{ X_i \}_{i = 1}^n$ are (sample) covariance matrices, where the underlying random vectors satisfy
appropriate tail or moment conditions.
\begin{corro}\label{corro:randomcov} Let $\pi_m$ be a probability distribution on $\R^m$ with second moment matrix $\Gamma \coloneq \E_{z \sim \pi_m}[z z^{\T}]$ satisfying $\lambda_{\min}(\Gamma) > 0$. Consider the random matrix ensemble
\begin{equation}\label{eq:samplecov}
\mc{M}(\pi_m, q) = \left \{\frac{1}{q} \sum_{k = 1}^q z_k z_k^{\T}, \; \, \{z_k \}_{k = 1}^q  \overset{\text{i.i.d.}}{\sim}
\pi_m \right \}.
\end{equation}
Suppose that $\{ X_i \}_{i = 1}^n \overset{\text{i.i.d.}}{\sim} \mc{M}(\pi_m, q)$ and let
$\wh{\Gamma}_n \coloneq \frac{1}{n} \sum_{i = 1}^n X_i$ and $0 < \epsilon_n < \lambda_{\min}(\Gamma)$. Under the event $\{\nnorm{\Gamma - \wh{\Gamma}_n}_{\infty} \leq \epsilon_n \}$, $\mc{X}$ satisfies Condition \ref{cond:slowrate} with
\begin{equation*}
R_* =  \frac{2 (\lambda_{\max}(\Gamma) + \epsilon_n)}{\lambda_{\min}(\Gamma) - \epsilon_n}  \quad \text{and} \; \; \, \, \tau_*^2 = (\lambda_{\max}(\Gamma) + \epsilon_n)^2.
\end{equation*}
\end{corro}
It is instructive to spell out Corollary \ref{corro:randomcov} with $\pi_m$ as the standard Gaussian distribution on $\R^m$. The
matrix $\wh{\Gamma}_n$ equals the sample covariance matrix computed from $N = n \cdot q$ samples. It is well-known (see e.g.~\cite{DavidsonSzarek}) that for $m,N$
large, $\lambda_{\max}(\wh{\Gamma}_n)$ and $\lambda_{\min}(\wh{\Gamma}_n)$ concentrate sharply around $(1 + \eta_n)^2$ and
$(1 - \eta_n)^2$, respectively, where $\eta_n = \sqrt{m/N}$. Hence, for any $\gamma > 0$, there exists $C_{\gamma} > 1$
so that if $N \geq C_{\gamma} m$, it holds that $R_* \leq 2 + \gamma$. Similar though weaker concentration results for
$\nnorm{\Gamma - \wh{\Gamma}_n}_{\infty}$ are available for the broad class of distributions $\pi_m$ having finite
fourth moments \cite{Vershynin2012}. When specialized to $q = 1$, Corollary \ref{corro:randomcov} yields a statement
about $\mc{X}$ made up from random rank-one measurements $X_i = z z^{\T}$, $i=1,\ldots,n$, cf.~\eqref{eq:rankone}. The preceding discussion
indicates that Condition \ref{cond:slowrate} tends to be satisfied in this case.

\textbf{Main result of this subsection.} We are now in position to state the following theorem.
\begin{theo}\label{theo:slowratebound} Suppose that model \eqref{eq:tracereg_spd} holds with $\mc{X}$ satisfying Condition \ref{cond:slowrate} with constants
$R_*$ and $\tau_*^2$. We then
have
\begin{align*}
\frac{1}{n} \nnorm{\mc{X}(\Sigma^*) - \mc{X}(\wh{\Sigma})}_2^2
%
&\leq \max \left\{2 (1 + R_*) \lambda_0 \nnorm{\Sigma^*}_1, \, 2 \lambda_0 \nnorm{\Sigma^*}_1 + 8 \left( \lambda_0 \frac{R_*}{\tau_*} \right)^2   \right \}
\end{align*}
where, for any $\mu \geq 0$, with probability at least $1 - (2m)^{-\mu}$
\begin{equation*}
\lambda_0 \leq \sigma \sqrt{(1 + \mu) 2 \log(2m)  \frac{V_n^2}{n}}, \quad \text{where} \; \; \, V_n^2 = \norm{\frac{1}{n} \sum_{i = 1}^n X_i^2}_{\infty}.
\end{equation*}
\end{theo}
\emph{Remarks.}
\begin{enumerate}
\item Under the scalings $R_* = O(1)$ and $\tau_*^2 = \Omega(1)$, the bound of Theorem \ref{theo:slowratebound} is
      of the order $O(\lambda_0 \nnorm{\Sigma^*}_1  + \lambda_0^2)$ as announced in \eqref{eq:slowrate_overview} at
      the beginning of this section.
\item For given $\mc{X}$, the quantity $\tau^2(\mc{X},R)$ can be evaluated by solving a least squares
      problem with \textsf{spd} constraints. Hence it is feasible to check in practice whether Condition
      \ref{cond:slowrate} holds. In fact, the bound of Theorem \ref{theo:slowratebound} can be replaced
      with
       \begin{equation*}
      \min_{R > 1} \max \left\{2 (1 + R) \lambda_0 \nnorm{\Sigma^*}_1, \, 2 \lambda_0 \nnorm{\Sigma^*}_1 + 8 \left( \lambda_0 \frac{R}{\tau(\mc{X}, R)} \right)^2   \right \}.
\end{equation*}
\item For later reference, it is of interest to evaluate the term $V_n^2$ for $\mc{M}(\pi_m, q)$ with $\pi_m$ as the
      standard Gaussian distribution. It is proved in Appendix \ref{sec:Vn} that with high probability, it holds that
      \begin{equation*}
      V_n^2 \leq \left(1 + q^{-1/2} + \sqrt{m/(nq)} \right)^2 \left(1 + \sqrt{m/q} + \sqrt{4(m/q) \log n} \right)^2 = O(m \log n)
      \end{equation*}
      as long as $m = O(nq)$.
\end{enumerate}

\subsection{Bound on the estimation error}

In the previous subsection, we did not make any assumptions about $\Sigma^*$ apart from $\Sigma^* \in \psd^m$. Henceforth,
we suppose that $\Sigma^*$ is of low rank $1 \leq r \ll m$ and study the performance of the constrained least squares
estimator \eqref{eq:constrainedleastsquares_compact} for prediction and estimation in such setting.

\textbf{Preliminaries.} Let $\Sigma^* = U \Lambda U^{\T}$  be the eigenvalue decomposition of $\Sigma^*$, where
\begin{equation*}
U = \left[ \begin{array}{cc}
                             U_{\pa} & U_{\perp} \\
                             \mbox{{\footnotesize $m \times r$}} & \mbox{{\footnotesize $m \times (m - r)$}}
                             \end{array} \right] \begin{bmatrix}
                             \Lambda_{r}   &  0_{r \times (m-r)}  \\
                                 0_{(m-r) \times r}        &  0_{(m - r) \times (m-r)}
                           \end{bmatrix}
                           \end{equation*}
where $\Lambda_r$ is diagonal with positive diagonal entries. Consider the linear subspace
\begin{equation*}
\TT^{\perp} = \{M \in \sym^m:\; M = U_{\perp} A U_{\perp}^{\T}, \quad A \in \sym^{m-r} \}.
\end{equation*}
From $U_{\pe}^{\T} \Sigma^* U_{\pe} = 0$, it follows that $\Sigma^*$ is contained in the orthogonal complement
\begin{equation*}
\TT = \{M \in \sym^m:\; M = U_{\pa} B + B^{\T} U_{\pa}^{\T}, \quad B \in \R^{r \times m} \},
\end{equation*}
which has dimension $m r - r(r - 1)/2 \ll \delta_m$ if $r \ll m$. The image
of $\TT$ under $\mc{X}$ is denoted by $\mc{T} = \{z \in \R^n:\; z = \mc{X}(M), \; \, M \in \TT \}$.

\textbf{Conditions on $\mc{X}$.} We now introduce the key quantities the bound in this subsection depends on.\\
\emph{Separability constant.}
\begin{align*}
\tau^2(\TT) &= \frac{1}{n} \text{dist}^2 \left(\mc{T}, \mc{P}_{\mc{X}} \right), \quad \; \mc{P}_{\mc{X}} \coloneq \{z \in \R^n:\; z = \mc{X}(\Delta), \; \Delta \in \TT^{\pe} \cap \mc{S}_1^+(m) \} \\
 &= \min_{\Theta \in \TT, \; \Lambda \in \mc{S}_1^+(m) \cap \TT^{\pe}} \frac{1}{n} \nnorm{\mc{X}(\Theta) - \mc{X}(\Lambda)}_2^2
\end{align*}
\emph{Restricted eigenvalue.}
\begin{equation*}
\phi^2(\TT) = \min_{0 \neq \Delta \in \TT} \, \frac{\nnorm{\mc{X}(\Delta)}_2^2/n}{\nnorm{\Delta}_1^2}.
\end{equation*}
As indicated by the following statement concerning the noiseless case, for bounding $\nnorm{\wh{\Sigma} - \Sigma^*}$, it is inevitable
to have lower bounds on the above two quantities.
\begin{prop}\label{prop:recovery_noiseless} Consider the trace regression model \eqref{eq:tracereg} with $\eps_i = 0$, $i=1,\ldots,n$.
Then
\begin{equation*}
\argmin_{\Sigma \in \psd^m} \frac{1}{2n} \nnorm{\mc{X}(\Sigma^*) - \mc{X}(\Sigma)}_2^2  = \{ \Sigma^* \} \; \; \; \text{for all} \; \; \Sigma^* \in \TT \cap \psd^m
\end{equation*}
if and only if it holds that $\tau^2(\TT) > 0$ and $\phi^2(\TT) > 0$.
\end{prop}
\emph{Correlation constant.} Moreover, we make use of the following the quantity. It is not yet clear to us
whether control of this quantity is intrinsically required, or whether its appearance in our bound is for merely
technical reasons.
\begin{equation*}
\mu(\TT) = \max \left\{\frac{1}{n}\scp{\mc{X}(\Delta)}{\mc{X}(\Delta')}:\;\,\nnorm{\Delta}_1\leq 1,\Delta \in \TT,
\; \,  \Delta' \in \mc{S}_1^+(m) \cap \TT^{\pe} \right \}.
\end{equation*}
We are now in position to provide a bound on $\nnorm{\wh{\Sigma} - \Sigma^*}_1$.
\begin{theo}\label{theo:estimationerror}
Suppose that model \eqref{eq:tracereg_spd} holds with $\Sigma^*$ as considered throughout this subsection and let $\lambda_0$ be
defined as in Theorem \ref{theo:slowratebound}. We then have
\begin{align*}
\nnorm{\wh{\Sigma} - \Sigma^*}_1 \leq \max &\Bigg\{8 \lambda_0 \frac{\mu(\TT)}{\tau^2(\TT) \phi^2(\TT)} \left(\frac{3}{2} + \frac{\mu(\TT)}{\phi^2(\TT)}   \right) + 4 \lambda_0 \left(\frac{1}{\phi^2(\TT)} + \frac{1}{\tau^2(\TT)} \right),  \\
&\frac{8 \lambda_0}{\phi^2(\TT)} \left( 1 + \frac{\mu(\TT)}{\phi^2(\TT)} \right), \; \frac{8 \lambda_0}{\tau^2(\TT)}   \Bigg \}.
\end{align*}
\end{theo}
\emph{Remark.} Given the above bound on $\nnorm{\wh{\Sigma} - \Sigma^*}_1$, it is possible to
obtain an improved bound on the prediction error scaling with $\lambda_0^2$ in place of $\lambda_0$,
cf.~\eqref{eq:basicinequality} in Appendix \ref{app:slowratebound}.\\
\\
The quality of the bound of Theorem \ref{theo:estimationerror} depends on how the quantities $\tau^2(\TT)$, $\phi^2(\TT)$ and $\mu(\TT)$ scale
with $n$, $m$ and $r$, which is highly design-dependent. Accordingly, the estimation error in nuclear norm can be
non-finite in the worst case and $O(\lambda_0 r)$ in the best case.
\begin{itemize}
\item The quantity $\tau^2(\TT)$ is specific to the geometry of the constrained least squares problem
      \eqref{eq:constrainedleastsquares_compact} and hence of critical importance.
For instance, it follows from Proposition \ref{prop:Amelunxen} that for standard Gaussian measurements, $\tau^2(\TT) = 0$
with high probability once $n < \frac{1}{2} \delta_m$. The situation can be much better for random
\textsf{spd} measurements \eqref{eq:samplecov} as exemplified for measurements $X_i = z_i z_i^{\T}$ with
$z_i \overset{\text{i.i.d.}}{\sim} N(0, I)$ in the subsequent section. Specifically, it turns out that $\tau^2(\TT) = \Omega(1/r)$ as long as
$n = \Omega(m \cdot r)$.
\item It is not restrictive to assume that the quantity $\phi^2(\TT)$ is positive. Indeed, without
      that assumption, even an oracle estimator based on knowledge of the subspace $\TT$ would
      fail. Reasonable sampling operators $\mc{X}$ have rank $\min\{n,\delta_m\}$ so that the nullspace of $\mc{X}$
      only has a trivial intersection with the subspace $\TT$ as long as $n \geq \dim(\TT) = mr - r(r-1)/2$.
\item For fixed $\TT$, computing $\mu(\TT)$ entails solving a biconvex (albeit non-convex) optimization problem in the variables
      $\Delta \in \TT$ and $\Delta' \in \mc{S}_1^+(m) \cap \TT^{\pe}$. Alternating optimization (also known
      as block coordinate descent) is a practical approach to such optimization problems for which a globally optimal solution is out of reach. In this manner
      we explore the scaling of $\mu(\TT)$ numerically as done for $\tau^2(\TT)$. We find that $\mu(\TT) = O(\delta_m/n)$ so that $\mu(\TT) = O(1)$ apart from the regime $n/\delta_m \rightarrow 0$, without ruling out the possibility of undersampling, i.e.~$n < \delta_m$.
\end{itemize}

\section{Numerical results}\label{sec:numerical}
In this section, we provide a series of empirical results regarding properties of the estimator $\wh{\Sigma}$.
In particular, its performance relative to regularization-based methods is explored. We also present an
application to spiked covariance estimation for the CBCL face image data set and stock prices from NASDAQ.

\subsection{Scaling of the constant $\tau^2(\TT)$}
For $\mc{X}$ and $\TT$ given, it is possible to evaluate $\tau^2(\TT)$ by solving a convex
optimization problem. This is different from other conditions employed in the literature
such as restricted strong convexity \cite{Negahban2011}, 1-RIP \cite{Chen2013} or
restricted uniform boundedness \cite{Cai2014} that involve a non-convex optimization problem
even for fixed $\TT$.

We here consider sampling operators with random i.i.d.~measurements $X_i = z_i z_i^{\T}$,
where $z_i \sim N(0, I)$ is a standard Gaussian random vector in $\R^m$ (equivalently,
$X_i$ follows a Wishart distribution) , $i=1,\ldots,n$. We expect $\tau^2(\TT)$ to behave
similarly for random rank-one measurements of the same form as long as the underlying
probability distribution has finite fourth moments, and thus for (a broad subclass of)
the ensemble $\mc{M}(\pi_m, q)$ \eqref{eq:samplecov}.

\begin{figure}
\begin{tabular}{ll}
\hspace*{-0.035\textwidth}\includegraphics[width = 0.48\textwidth]{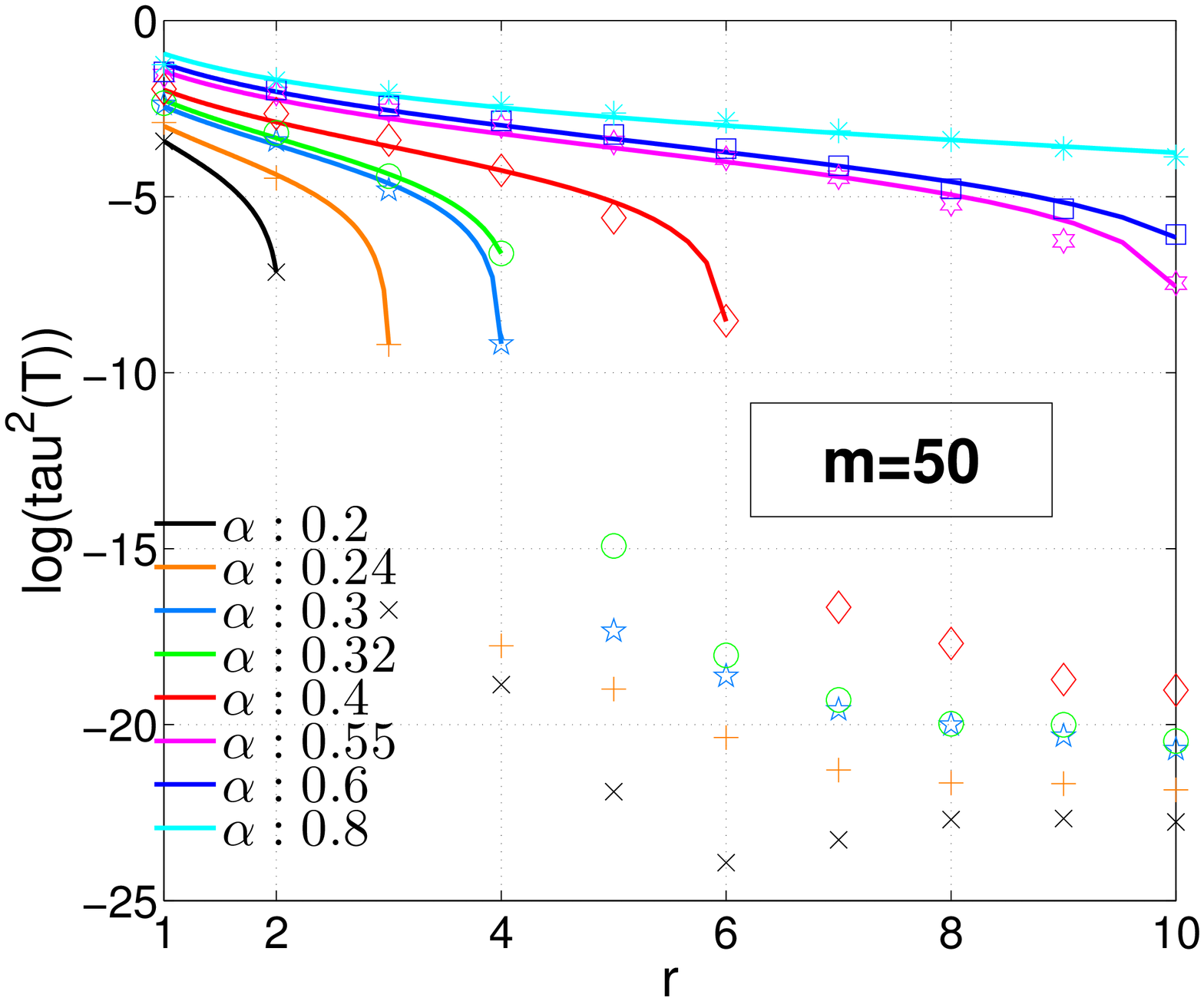} & \includegraphics[width = 0.48\textwidth]{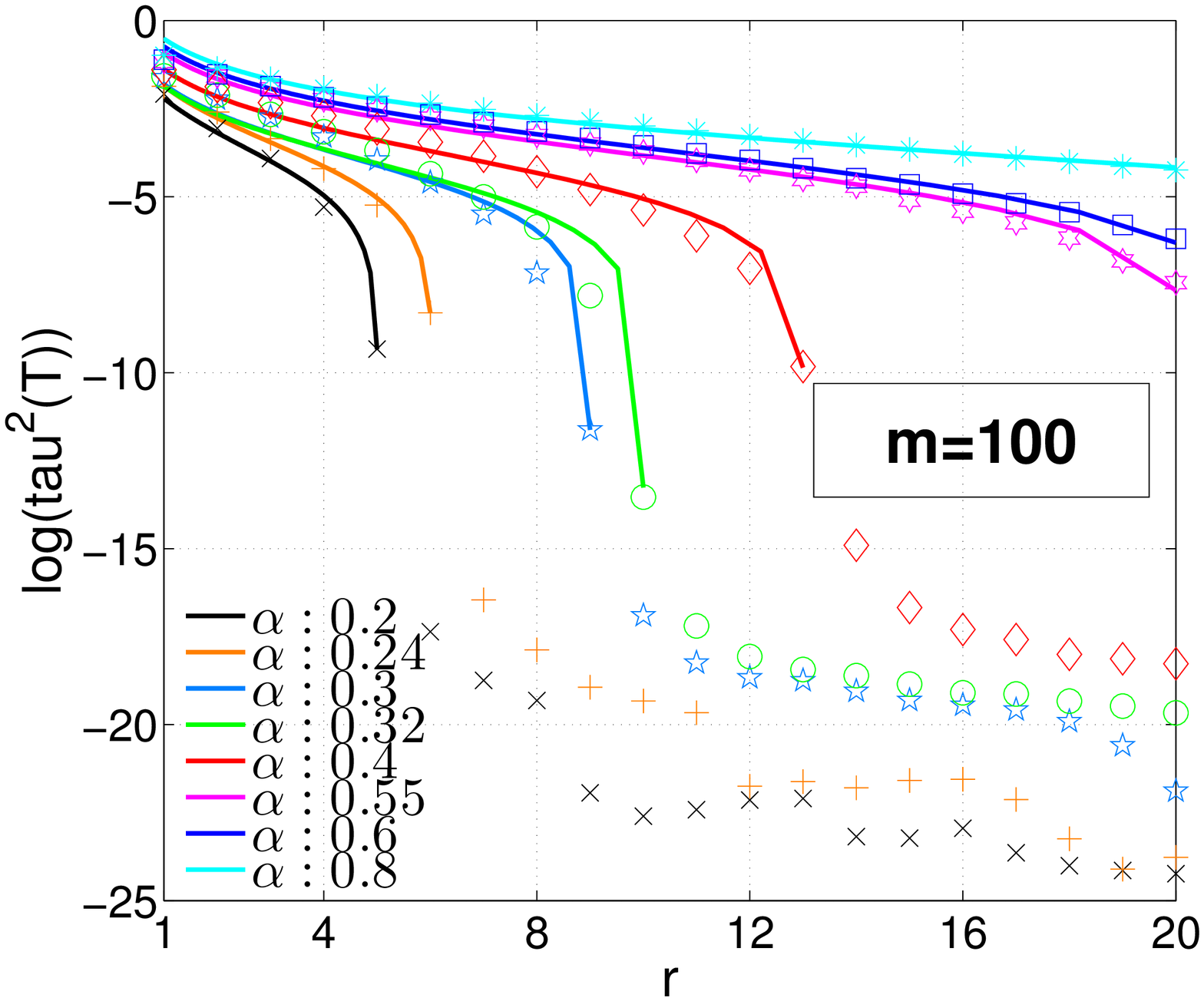}
\end{tabular}
\caption{Scaling of $\log \tau^2(\TT)$ in dependence of $r$ (horizontal axis) and $\alpha = n/\delta_m$ (colors/symbols). The
solid lines represent the fit of model \eqref{eq:model_tauTT}. Note that the curves are only fitted to those points for which
$\tau^2(\TT)$ exceeds $10^{-6}$. Best seen in color.}\label{fig:tauT}
\end{figure}

In order to explore the scaling of $\tau^2(\TT)$ with $n$, $m$ and $r$, we fix
$m \in \{30,50,70,100 \}$. For each choice of $m$, we vary $n = \alpha \delta_m$, where a grid
of $20$ values ranging from $0.16$ to $1.1$ is considered $\alpha$. For $r$, we consider the grid $\{1,2,\ldots,m/5\}$.
For each combination of $m$, $n$, and $r$, we use 50 replications. Within each replication, the subspace
$\TT$ is generated randomly from the eigenspace associated with the non-zero eigenvalues of a random matrix
$G^{\T} G$, where the entries of the $m \times r$ matrix $G$ are i.i.d.~$N(0,1)$.\\
The results point to the existence of a phase transition as it is typical for problems related to that
under study \cite{Amelunxen2014}. Specifically, it turns out that the scaling of $\tau^2(\TT)$ can be
well described by the relation
\begin{equation}\label{eq:model_tauTT}
\tau^2(\TT) \approx \phi_{m,n} \max\{1/r - \theta_{m,n}, 0\},
\end{equation}
where $\phi_{m,n}, \theta_{m,n} > 0$ depend on $m$ and $n$. In order to arrive at model
\eqref{eq:model_tauTT}, we first obtain the $5$\%-quantile as summary statistic of the 50 replications
associated with each triple $(n,m,r)$. At this point, note that the use of the mean as a summary statistic is
not appropriate as it may mask the fact that the majority of the observations are zero. For each pair of $(n,m)$,
we then identify all values of $r$ for which the corresponding $5$\%-quantile drops below $10^{-6}$, which serves
as effective zero here. For the remaining values, we fit model \eqref{eq:model_tauTT} using nonlinear least squares
(working on a log scale). Figure \ref{fig:tauT} shows that model \eqref{eq:model_tauTT} provides a rather accurate
description of the given data. Concerning $\phi_{m,n}$ and $\theta_{m,n}$, the scalings $\phi_{m,n} = \phi_0 n/m$ and
$\theta_{m,n} = \theta_0 m/n$ for constants $\phi_0, \theta_0 > 0$ appear to be reasonable. This gives rise to the
requirement $n > \theta_0 (m r)$ for exact recovery to be possible in the noiseless case
(cf.~Proposition \ref{prop:recovery_noiseless}) and yields that $\tau^2(\TT) = \Omega(1/r)$ as long as $n = \Omega(mr)$,

\subsection{Comparison with regularization-based approaches}

In this subsection, we empirically evaluate $\nnorm{\wh{\Sigma} - \Sigma^*}_1$ relative
to regularization-based methods proposed in the literature.\\

\textbf{Setup.} We consider Wishart measurement matrices as in the previous subsection. Again,
we expect a similar behaviour for (most) other random designs from ensemble $\mc{M}(\pi_m, q)$. We
fix $m = 50$ and let $n \in \{0.24, 0.26, \ldots, 0.36, 0.4, \ldots, 0.56 \} \cdot m^2$ and $r \in \{1,2,\ldots,10 \}$ vary.
For each configuration of $n$ and $r$, we consider 50 replications. In each of these replications, we generate data
\begin{equation}\label{eq:model_experiments}
y_i = \tr(X_i \Sigma^*) + \sigma \eps_i, \; \, \sigma = 0.1, \; \; \, i=1,\ldots,n,
\end{equation}
where $\Sigma^*$ is generated as the sum of $r$ Wishart matrices and the $\{\eps_i \}_{i = 1}^n$ are i.i.d.~$N(0,1)$.
\begin{figure}[ht!]
\begin{tabular}{lll}
\includegraphics[width = 0.48\textwidth]{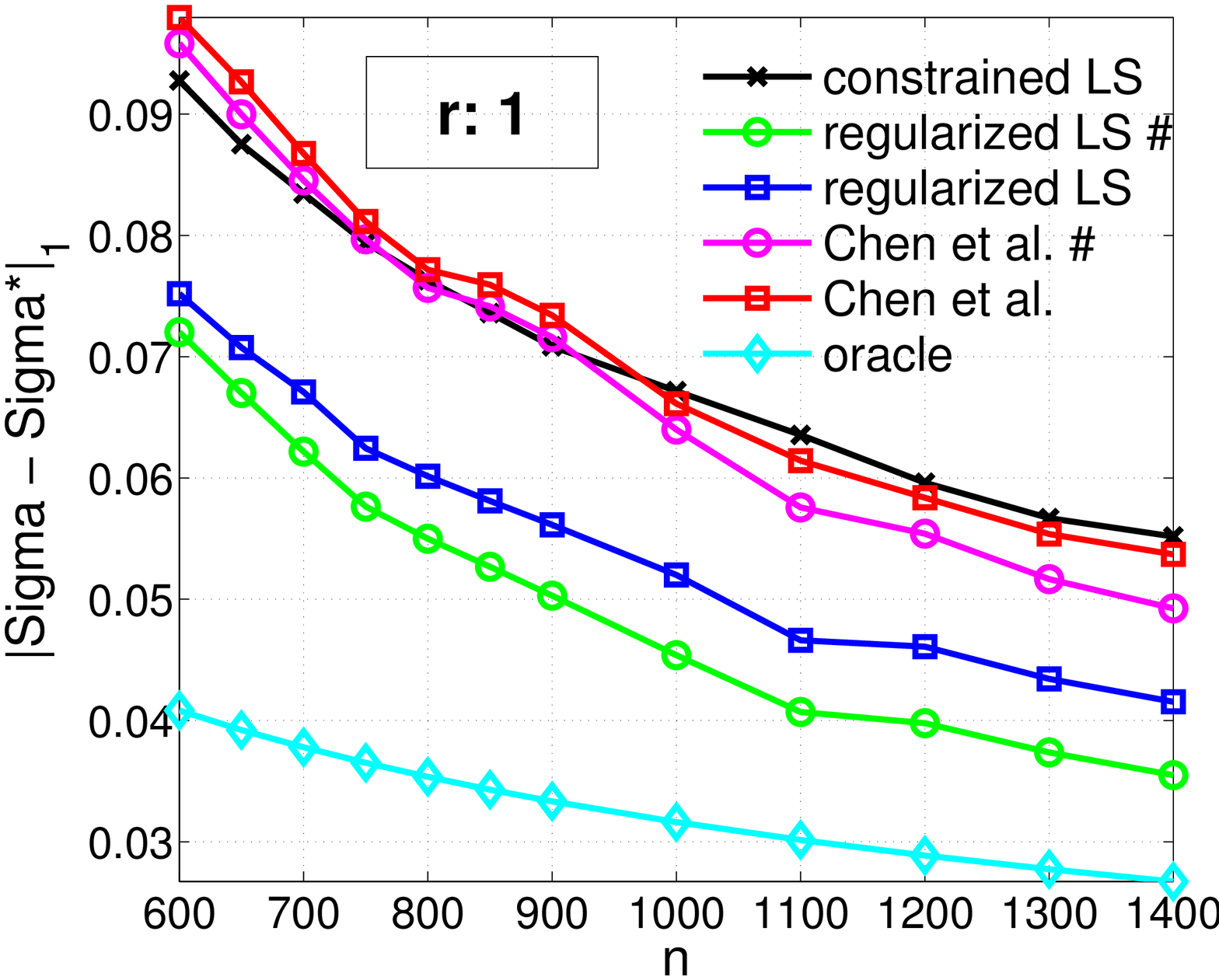} &
\includegraphics[width = 0.48\textwidth]{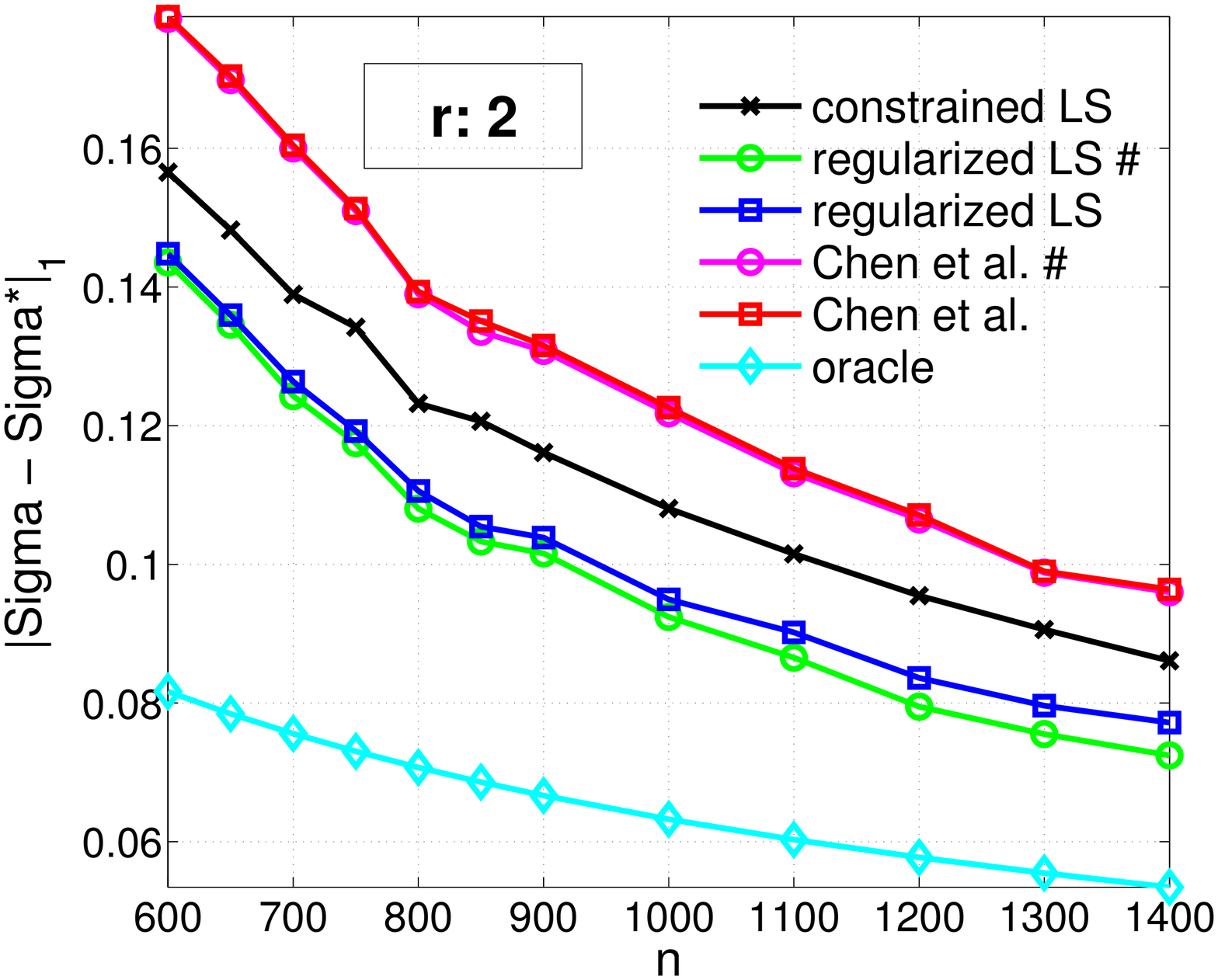} \\
\includegraphics[width = 0.48\textwidth]{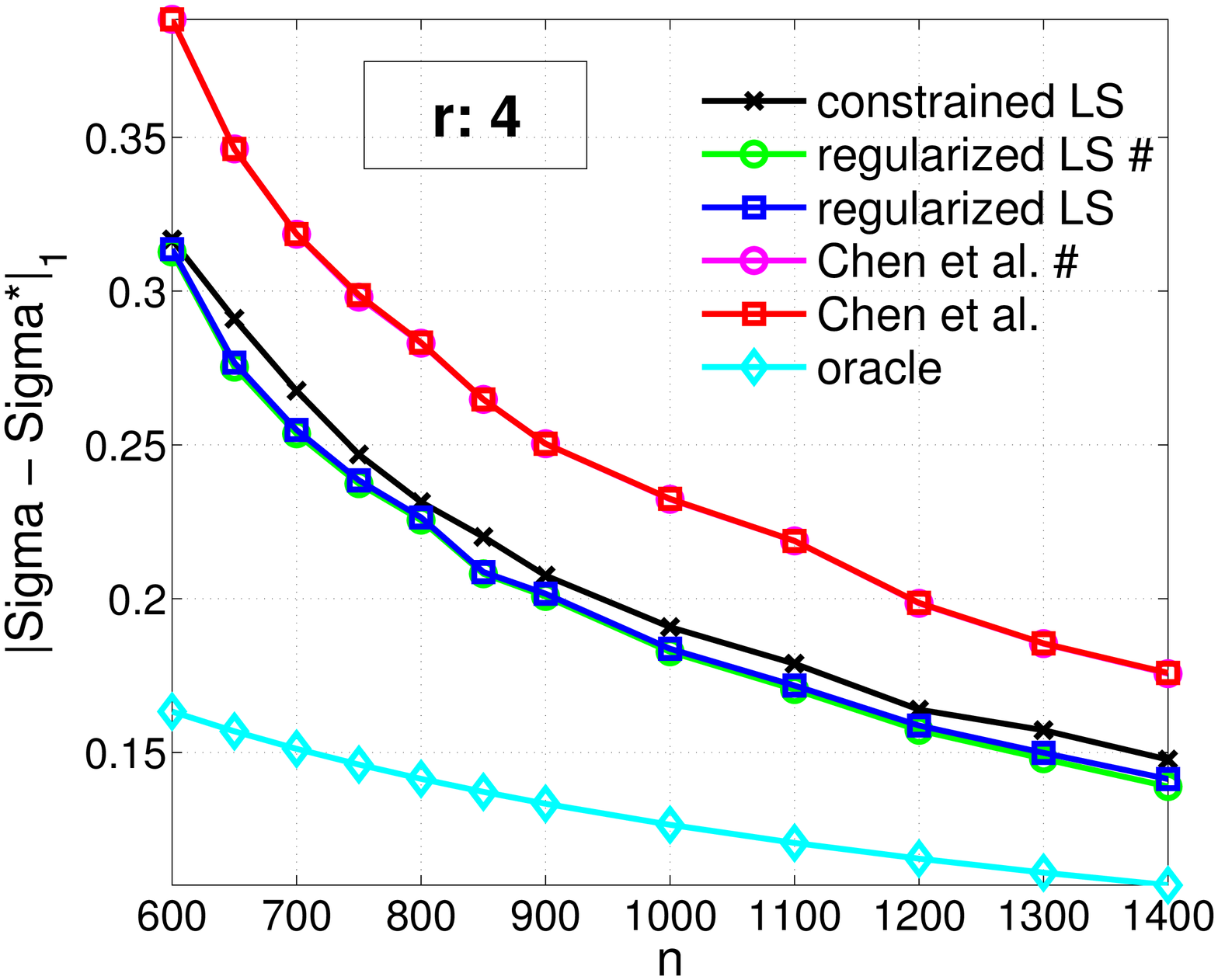} &
\includegraphics[width = 0.48\textwidth]{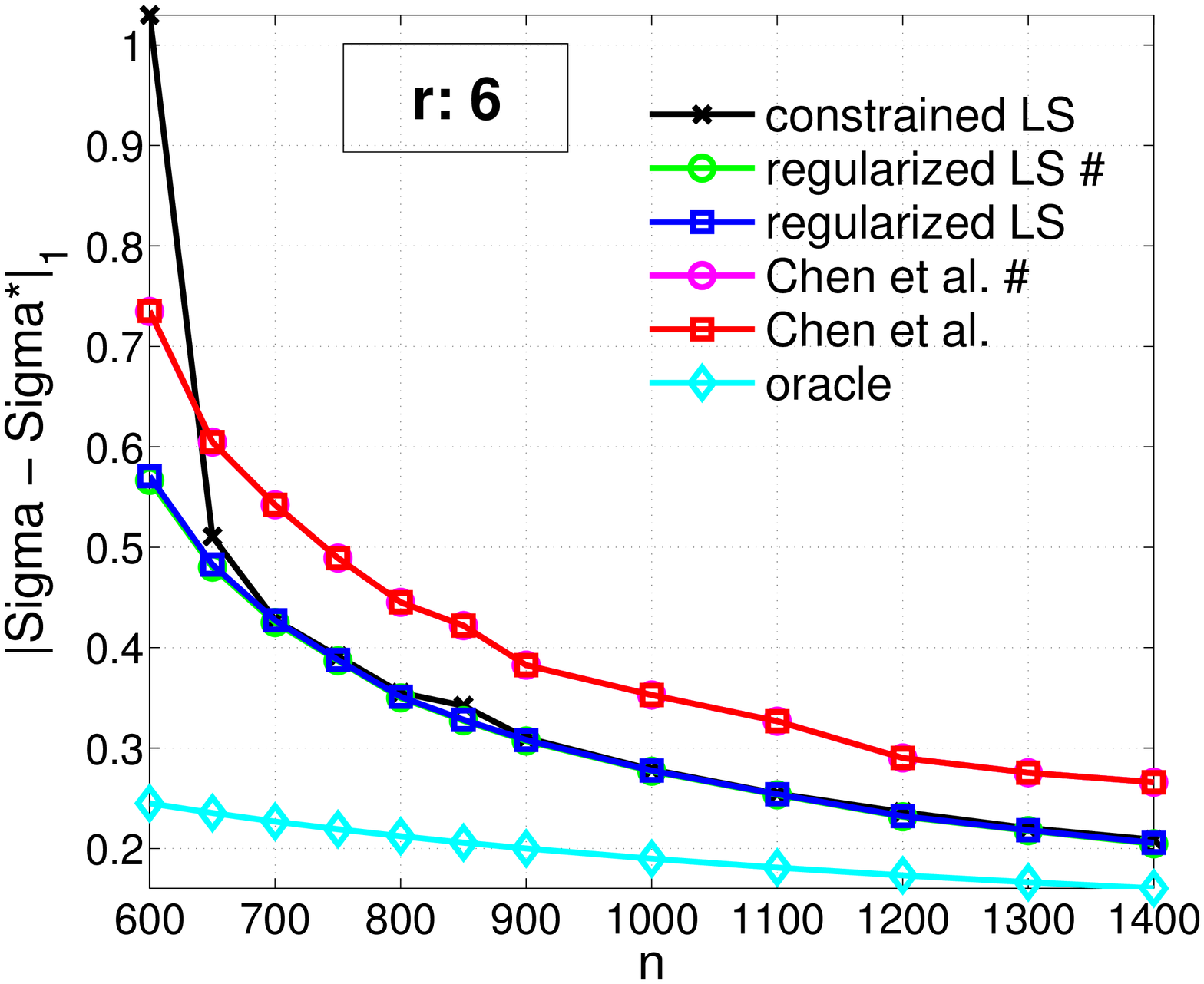} \\
\includegraphics[width = 0.48\textwidth]{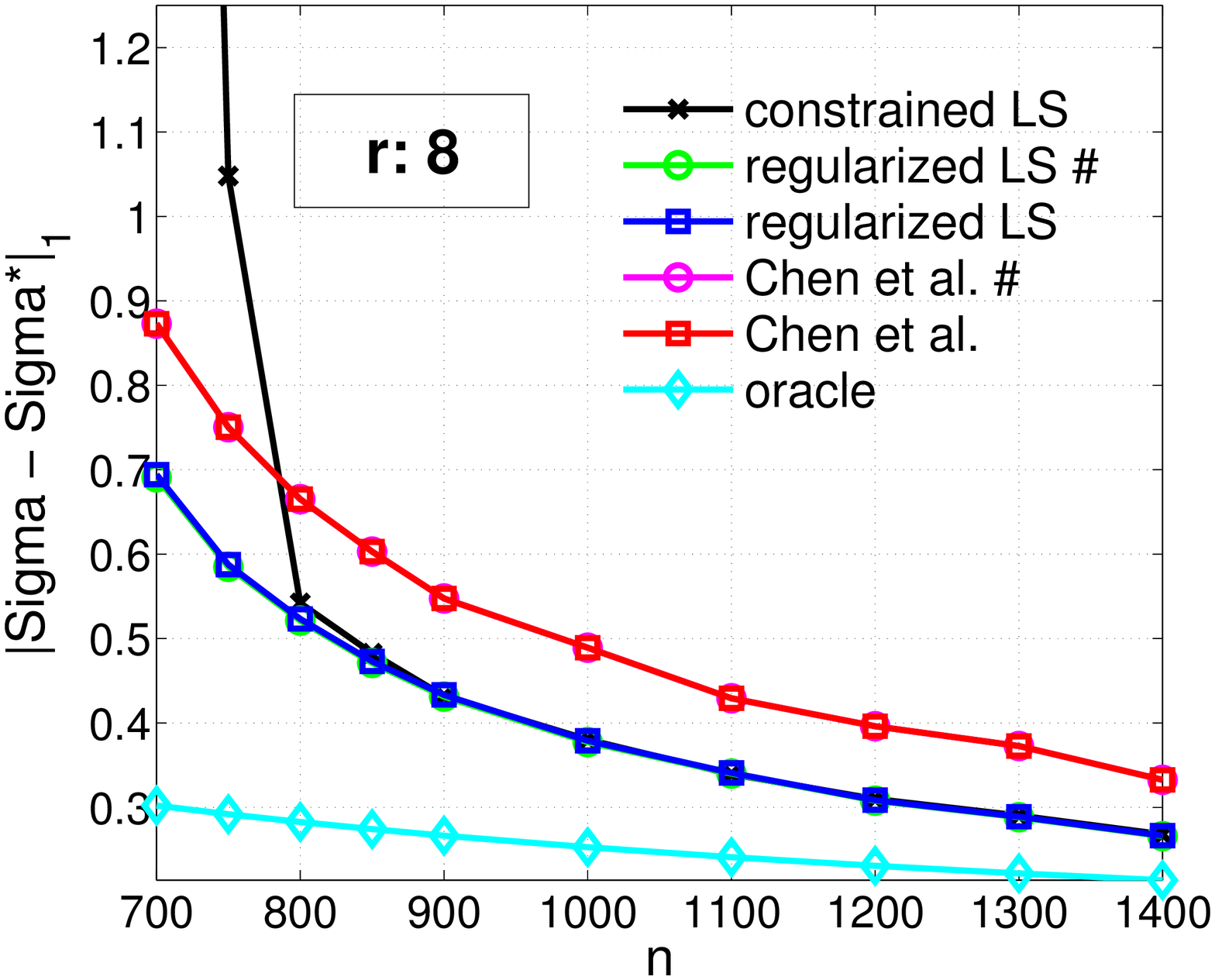} &
\includegraphics[width = 0.48\textwidth]{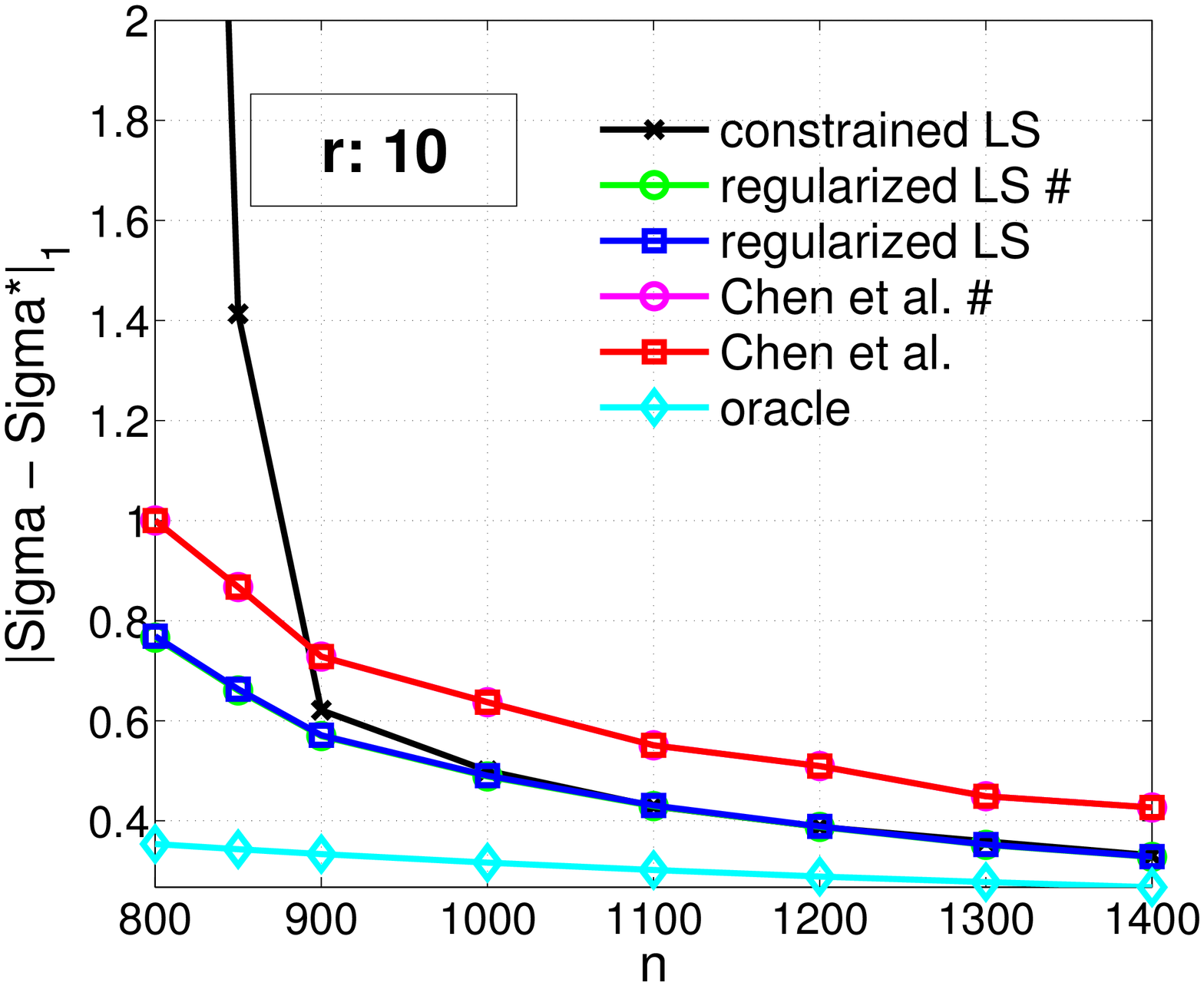}
\end{tabular}
\caption{Average estimation error (over 50 replications) in nuclear norm for fixed $m = 50$ and certain choices of $n$ and $r$.
In the legend, ``LS'' is used as a shortcut for ``least squares''. Chen et al.~refers to \eqref{eq:constrainednuclearnorm}. ``\#''indicates an oracular choice
of the tuning parameter. ``oracle'' refers to the ideal error $\sigma r \sqrt{m/n}$. Standard error bars are not shown
as the standard errors are of negligible magnitude. Best seen in color.}\label{fig:ell1errors}
\end{figure}

\textbf{Regularization-based approaches.} We compare $\wh{\Sigma}$ to the corresponding nuclear norm
regularized estimator in \eqref{eq:nnls_tracenormreg}. Regarding the choice of the regularization parameter $\lambda$, we consider the grid $\lambda_* \cdot \{0.01, 0.05, 0.1, 0.3, 0.5, 1, 2, 4, 8, 16\}$, where
$\lambda_* = \sigma \sqrt{m/n}$ as recommended in \cite{Negahban2011} and pick $\lambda$ so that the prediction
error on a separate validation data set of size $n$ generated from \eqref{eq:model_experiments} is minimized.
Note that in general, neither $\sigma$ is known nor an extra validation data set is available. Our goal here
is to ensure that the regularization parameter is properly tuned. In addition, we consider an oracular
choice of $\lambda$ where $\lambda$ is picked from the above grid such that the performance measure of interest (the distance
to the target in the nuclear norm) is minimized. We also compare to the constrained nuclear norm minimization approach
of Chen et al.~\cite{Chen2013} given by
\begin{equation}\label{eq:constrainednuclearnorm}
\min_{\Sigma} \tr(\Sigma) \quad \text{subject to} \; \; \, \Sigma \gec 0, \; \;  \text{and} \; \; \nnorm{y - \mc{X}(\Sigma)}_1 \leq \lambda.
\end{equation}
For the parameter $\lambda$, we consider the grid $n \sigma \sqrt{2/\pi} \cdot \{0.2, 0.3, \ldots, 1, 1.25\}$. This specific
choice is motivated by the observation that $\E[\nnorm{y - \mc{X}(\Sigma^*)}_1] = \E[\nnorm{\eps}_1] = n \sigma \sqrt{2/\pi}$.
Apart from that, tuning of $\lambda$ is performed as for the nuclear norm regularized estimator. In addition, we have
assessed the performance of the approach in \cite{Cai2014}, which does not impose an \textsf{spd} constraint but
adds one more constraint to the formulation \eqref{eq:constrainednuclearnorm}. That additional constraint significantly
complicates optimization of the problem and yields a second tuning parameter. Therefore, instead of doing a grid search over a 2D-grid, we use fixed values as specified in \cite{Cai2014} given the knowledge of $\sigma$. The results are similar or worse than those of \eqref{eq:constrainednuclearnorm} (note in particular that positive semidefiniteness is not taken advantage of in the approach of \cite{Cai2014}) and are hence not
reported here.

\textbf{Discussion of the results.} We can conclude from Figure \ref{fig:ell1errors} that in most cases,
the performance of the constrained least squares estimator does not differ much from that of the regularization-based
methods with careful parameter tuning, which are not too far from the oracle. However, for larger values of $r$, the constrained least squares
estimator seems to require slightly more measurements to achieve competitive performance.
\subsection{Real data examples}

We conclude this section by presenting an application to recovery of spiked covariance matrices, a notion
due to \cite{Johnstone2001}.

\textbf{Background.}
A spiked covariance matrix is of the form $\Sigma^* = \sum_{j = 1}^r \lambda_j u_j u_j^{\T} + \sigma^2 I$, where $r \ll m$ and
$\lambda_j \gg \sigma^2 > 0$, $j=1,\ldots,r$. Note that for data $\{ z_i \}_{i = 1}^n$ following the factor model
\begin{equation}\label{eq:factormodel}
z_i  = \sum_{j = 1}^r \alpha_{ij} f_j + \sigma \xi_i, \quad \xi_i \sim N(0, I),
\end{equation}
for orthogonal factors $\{ f_j \}_{j = 1}^r$ and random coefficients $\alpha_{ij} \sim N(0, \lambda_j)$ independent
from $\xi_i$, the population covariance matrix $\E[z_i z_i^{\T}]$, $i=1,\ldots,n$, is of the form given above. Model
\eqref{eq:factormodel} is one possible way to motivate principal components analysis (PCA); this connection explains
the relevance and the popularity of spiked covariance models.

\textbf{Extension to the spiked case.}
So far, we have assumed that the target $\Sigma^*$ is of low rank, but it is straightforward to extend the proposed
approach to the case in which $\Sigma^*$ is spiked as long as $\sigma^2$ is known or an estimate is available. A constrained least squares   estimator of $\Sigma^*$ takes the form $\wh{\Sigma} + \sigma^2 I$, where
\begin{equation}\label{eq:constrainedleastsquares_spiked}
\wh{\Sigma} \in \argmin_{\Sigma \in \psd^m} \frac{1}{2n} \nnorm{y - \mc{X}(\Sigma + \sigma^2 I)}_2^2.
\end{equation}
\textbf{Data sets.} (1) The CBCL facial image data set \cite{CBCL} consist of $N = 2429$ images of $19 \times 19$ pixels (i.e., ~$m = 361$). We take $\Sigma^*$ as the sample covariance matrix of this data set. It turns out that $\Sigma^*$ can be well
approximated by $\Sigma_r$, $r = 50$, where $\Sigma_r$ is the best rank $r$ approximation to $\Sigma^*$ obtained from computing
its eigendecomposition and setting to zero all but the top $r$ eigenvalues. (2) We construct a second data set from the daily end
prices of $m = 252$ stocks from the technology sector in NASDAQ, starting from the beginning of the year 2000 to the end of the year 2014
(in total $N = 3773$ days, retrieved from \texttt{finance.yahoo.com}). We take $\Sigma^*$ as the resulting sampling correlation matrix and choose
$r = 100$.

\textbf{Experimental setup.} As in all preceding measurements, we consider $n$ random Wishart measurements for the operator
$\mc{X}$, where $n = C (m r)$, where $C$ ranges from $0.25$ to $12$. Since
$\nnorm{\Sigma_r - \Sigma^*}_F / \nnorm{\Sigma^*}_F \approx 10^{-3}$ for both data sets, we work with $\sigma^2 = 0$ in \eqref{eq:constrainedleastsquares_spiked} for simplicity. To make the problem of recovering $\Sigma^*$ more difficult, we introduce additional noise to the problem by
using observations
\begin{equation}\label{eq:subsampling}
y_i = \tr(X_i S_i), \quad i=1,\ldots,n,
\end{equation}
where $S_i$ is an approximation to $\Sigma^*$ obtained from the sample covariance respectively sample correlation matrix of
$\beta N$ data points randomly sampled with replacement from the entire data set, $i=1,\ldots,n$,
where $\beta$ ranges from $0.4$ to $1/N$ ($S_i$ is computed from a single data point). For each choice of $n$ and $\beta$, $20$
replications are considered. The reported results are averages over these replications.

\begin{figure}
\subfigure{	
	\includegraphics[width=0.48\textwidth]{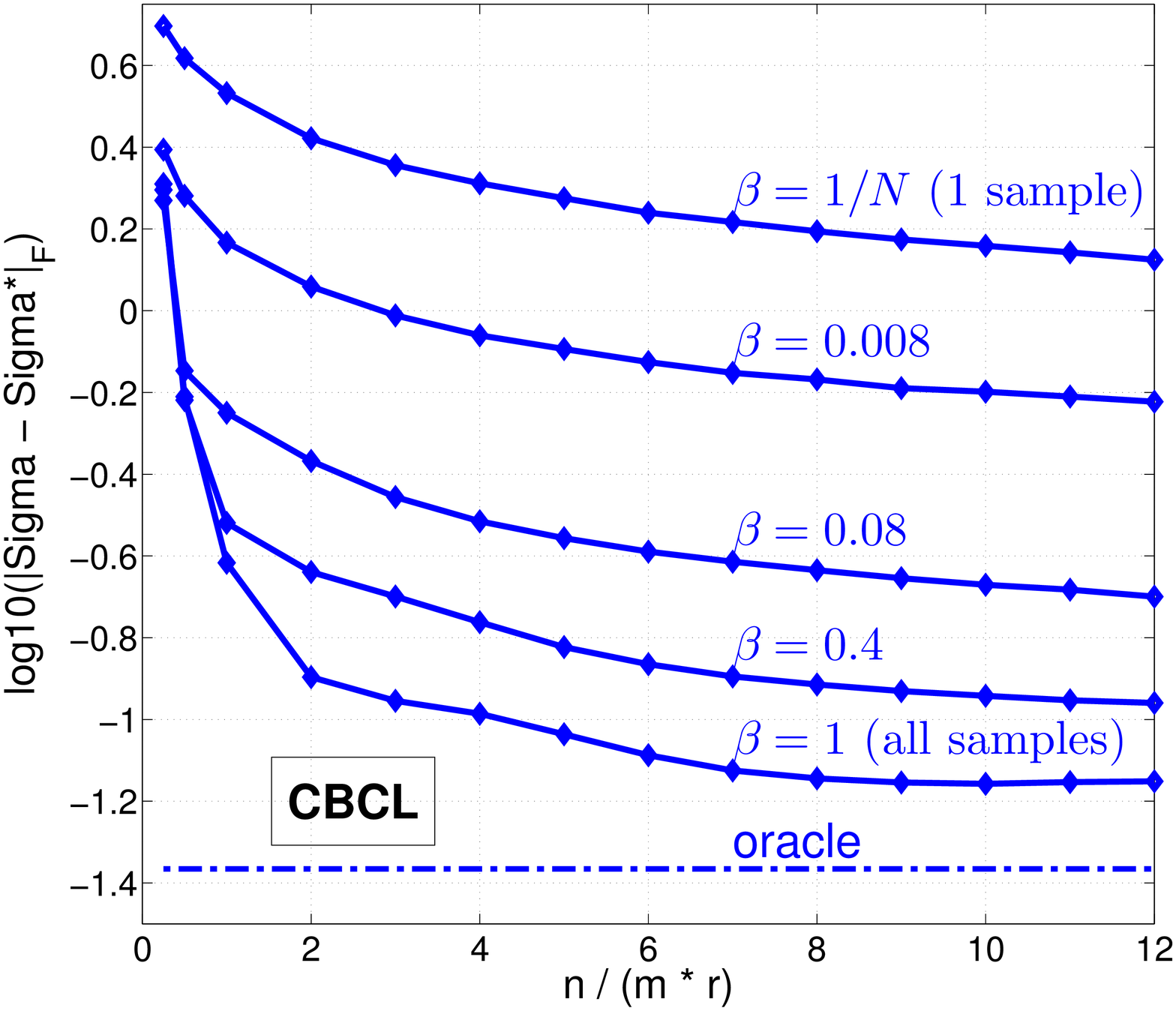}}
\subfigure{
	\includegraphics[width=0.48\textwidth]{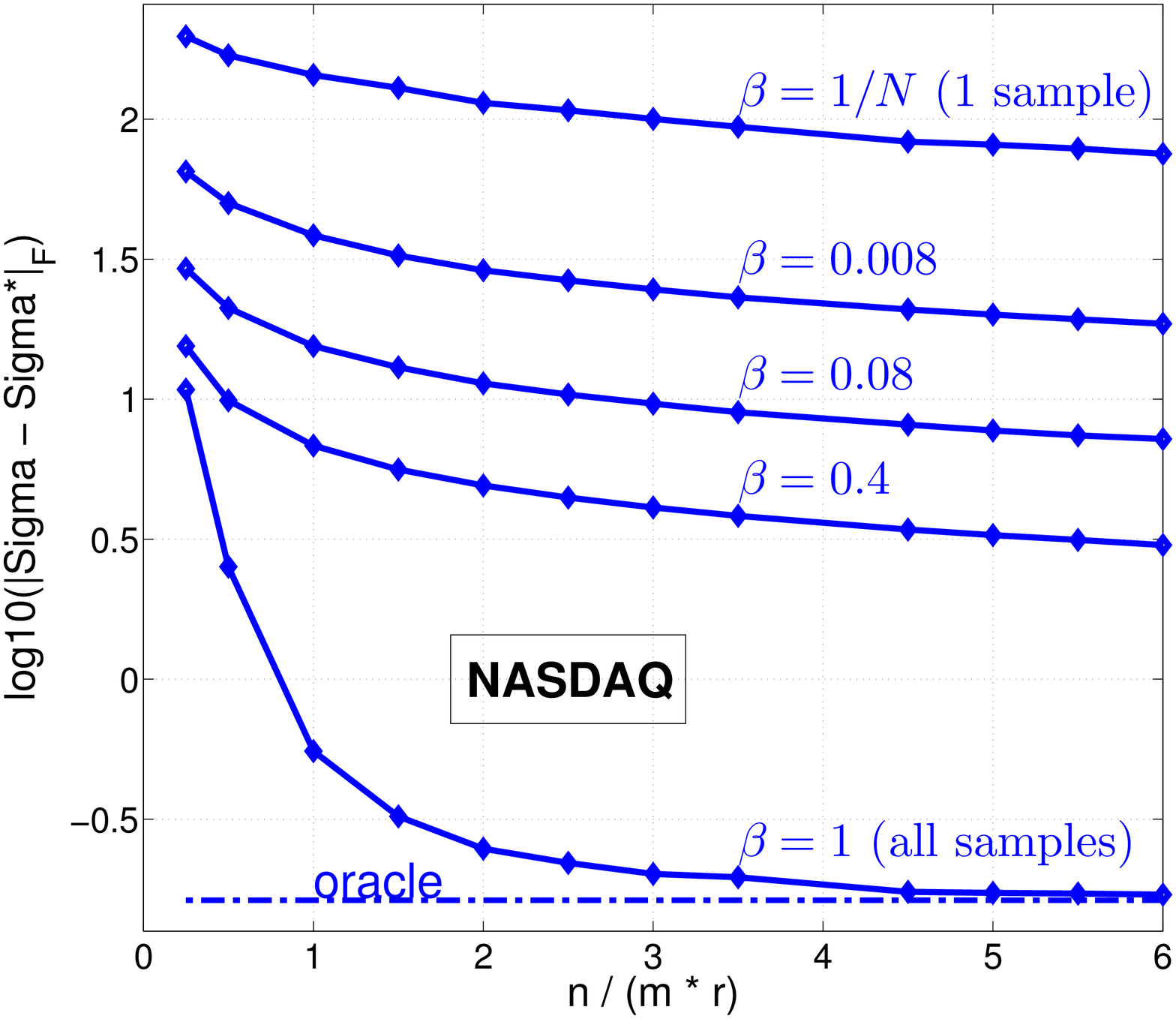}
}		
\caption{Average reconstruction errors $\log_{10} \nnorm{\wh{\Sigma} - \Sigma^*}_F$ in dependence of
$n/(mr)$ and the parameter $\beta$. ``oracle'' refers
to the best rank $r$-approximation $\Sigma_{r}$. Standard errors are one order of magnitude
smaller, and are thus omitted.}
\label{fig:errs_cbcl_nasdaq}
\end{figure}

\textbf{Results.} For the CBCL data set, it can be seen from Figure \ref{fig:errs_cbcl_nasdaq} and Table \ref{tab:errs_cbcl_nasdaq},
that $\wh{\Sigma}$ accurately approximates
$\Sigma^*$ (within a factor of three of the best rank-$r$ approximation $\Sigma_r$) once the number of measurements crosses
$2mr$. Performance degrades once additional noise is introduced to the problem by using measurements \eqref{eq:subsampling} that
are taken from a perturbed version of $\Sigma^*$. Even under significant perturbations ($\beta = 0.08$), reasonable reconstruction
of $\Sigma^*$ remains possible, albeit the number of required measurements increases accordingly. In the extreme case $\beta = 1/N$,
the error is still decreasing with $n$, but millions of samples seems to be required to achieve reasonable reconstruction error
(for computational reasons, we stop at $n = 12 m r \approx 216,000$).\\
The general picture is similar for the NASDAQ data set, but the difference between using measurements based on the
full sample correlation matrix on the one hand and approximations based on random subsampling \eqref{eq:subsampling} on the other hand are more pronounced. For $\beta = 1$, the reduction in error with increasing $n$ progresses visibly faster as for the first data set, and a
smaller error relative to $\Sigma_r$ close to $1$ is achieved.

\begin{table}
\begin{minipage}{\textwidth}
\hspace{0.07\textwidth}\textbf{CBCL} \hspace{0.4\textwidth} \textbf{NASDAQ} \\[1ex]
\begin{tabular}{|c|c|c|c|c|c|}
\hline
$\beta$ & {\small 1} & {\small 1} & {\small .4} & {\small .4} & {\small .08}  \\
\hline
$C$     &  {\small 2} & {\small 6} & {\small 4} & {\small 6} & {\small 10}  \\
\hline
\rule{0pt}{3ex}
$\frac{\nnorm{\wh{\Sigma} - \Sigma^*}_F}{\nnorm{\Sigma_r - \Sigma^*}_F}$  & {\small $<3$} & {\small $<2$} & {\small $~4$}  & {\small $~3$}  &  {\small $~5$} \\
\hline
\end{tabular}
\hspace{0.025\textwidth}
\begin{tabular}{|c|c|c|c|c|c|}
\hline
$\beta$ & {\small 1} & {\small 1} & {\small 1} & {\small 1} \\
\hline
$C$     &  {\small 1} & {\small 2} & {\small 3} & {\small 6} \\
\hline
\rule{0pt}{3ex}
$\frac{\nnorm{\wh{\Sigma} - \Sigma^*}_F}{\nnorm{\Sigma_r - \Sigma^*}_F}$  & {\small $<3.5$} & {\small $<2$} & {\small $<1.3$}  & {\small $<1.1$}  \\
\hline
\end{tabular}
\end{minipage}
\caption{Average reconstruction errors relative to $\Sigma_r$ for some selected values of $\beta$ and $n/(mr)$.}
\label{tab:errs_cbcl_nasdaq}
\end{table}

\section{Conclusion}\label{sec:summary}
In this paper, we have investigated trace regression
in the situation that the underlying matrix is symmetric positive semidefinite. We have
shown that under certain restrictions on the design, the constrained least squares
estimator enjoys excellent statistical properties similar to methods employing
nuclear norm regularization. This may come as a surprise, as regularization is
widely regarded as necessary in small sample settings. On the application side,
we have pointed out the usefulness of our findings for recovering spiked
covariance matrices from quadratic measurements.

\section*{Acknowledgement}
The work of Martin Slawski and Ping Li is partially supported by NSF-DMS-1444124, NSF-III-1360971, ONR-N00014-13-1-0764, and AFOSR-FA9550-13-1-0137.



\appendix

\section{Proof of Proposition \ref{prop:orthonormal}}

By rotational invariance of the Gaussian distribution of $\eps$, it suffices to consider the canonical orthonormal basis
of $\sym^m$ given by
\begin{align*}
&X_1 = e_1 e_1^{\T}, \; \, X_2 = \frac{1}{\sqrt{2}} (e_1 e_2^{\T} + e_2 e_1^{\T}), \ldots,
X_m = \frac{1}{\sqrt{2}} (e_1 e_m^{\T} + e_m e_1^{\T}), \; \, X_{m+1} =  e_2 e_2^{\T}, \\
&X_{m+2} = \frac{1}{\sqrt{2}} (e_2 e_3^{\T} + e_3 e_2^{\T}), \ldots, X_{\delta_m - 1} = \frac{1}{\sqrt{2}} (e_{m-1} e_{m}^{\T} + e_{m} e_{m-1}^{\T}), \; \, X_{\delta_m} = e_m e_m^{\T},
\end{align*}
where $\{ e_j \}_{j = 1}^m$ denote the canonical basis vectors of $\R^m$.
Equivalently, the corresponding map $\mc{X}: \sym^m \rightarrow \R^{\delta_m}$ equals the symmetric vectorization operator
\begin{equation}\label{eq:svecSigma}
\Sigma = (\sigma_{jk}) \mapsto (\sigma_{11}, \sqrt{2} \sigma_{12}, \ldots, \sqrt{2} \sigma_{1m}, \sigma_{22}, \sqrt{2} \sigma_{23}, \ldots,
\sqrt{2} \sigma_{(m-1) m}, \sigma_{mm})^{\T}
\end{equation}
Accordingly, denote by $\{ \eps_{jk} \}_{1 \leq j \leq k \leq m}$ the error terms corresponding to the entries
$\{ \sigma_{jk} \}_{1 \leq j \leq k \leq m}$. The minimization problem \eqref{eq:constrainedleastsquares_compact} can hence be expressed as
\begin{align}
&\min_{\Sigma \in \psd^m} \frac{1}{2n} \left\{ \sum_{j = 1}^m (\eps_{jj} - \sigma_{jj})^2 + \sum_{j < k} (\eps_{jk} - \sqrt{2} \sigma_{jk})^2 \right \} \notag \\
&= \min_{\Sigma \in \psd^m} \frac{1}{2n} \left\{ \sum_{j = 1}^m (\eps_{jj} - \sigma_{jj})^2 + 2 \sum_{j < k} \left(\frac{\eps_{jk}}{\sqrt{2}} - \sigma_{jk} \right)^2
\right \} \notag \\
&= \min_{\Sigma \in \psd^m} \nnorm{E - \Sigma}_F^2 \label{eq:projectiononpsd},
\end{align}
where the matrix $E = \mc{X}^*(\eps)$ has entries $E_{jj} = \eps_{jj}$, $j=1,\ldots,m$, and $E_{jk} = \eps_{jk}/\sqrt{2}$, $j,k=1,\ldots,m, \; j \neq k$. Now observe that the minimizer $\wh{\Sigma}$ of \eqref{eq:projectiononpsd} coincides with
the Euclidean projection of $E$ on $\psd^m$. It is well-known \cite{BoydVandenberghe2004} that the projection of a symmetric
matrix on the positive semidefinite cone is obtained by setting all its negative eigenvalues to zero, i.e., ~in terms of the eigendecomposition of $E = \sum_{j = 1}^p \lambda_j(E) u_j^{\T} u_j^{\T}$, we have
\begin{equation*}
\wh{\Sigma} = \sum_{j = 1}^m \max \{\lambda_j(E), 0 \} u_j u_j^{\T}.
\end{equation*}
At this point, we note that $E$ is a Wigner matrix, whose empirical distribution of its eigenvalues follows
Wigner's semicircle law as $m \rightarrow \infty$ (cf.~\cite{Tao2012}), which is symmetric around zero. Consequently,
we have
\begin{equation*}
\nnorm{\mc{X}(\wh{\Sigma})}_2^2 = \nnorm{\wh{\Sigma}}_F^2 = \sum_{j = 1}^m \{\lambda_j(E), 0 \}^2 \rightarrow \frac{1}{2}\nnorm{E}_F^2 \rightarrow \frac{\sigma^2}{2} \delta_m   \; \, \text{in probability as} \; m \rightarrow \infty.
\end{equation*}

\section{Proof of Proposition \ref{prop:Amelunxen}}

The proof of Proposition \ref{prop:Amelunxen} follows from results in \cite{Amelunxen2014}.
\begin{defnApp} Let $\mc{C} \subseteq \R^d$ be a convex cone. The statistical dimension
of $\mc{C}$ is defined as $\delta(\mc{C}) = \E[\nnorm{\Pi_{\mc{C}} g}_2^2]$, where $\Pi_{\mc{C}}$ denotes
the Euclidean projection onto $\mc{C}$ and the entries of $g$ are  i.i.d.~$N(0,1)$.
\end{defnApp}
\begin{theoremApp}\label{theo:amelunxenmain}\cite{Amelunxen2014} Let $f:\R^d \rightarrow \R \cup \{-\infty,+\infty\}$ be
a proper convex function. Suppose that $A \in \R^{n \times d}$ has i.i.d.~$N(0,1)$ entries, and let $z_0 = A x_0$ for a fixed $x_0 \in \R^d$. Consider the convex optimization problem
\begin{equation}\label{eq:randomconvexprog}
\text{minimize} \; f(x) \quad \text{subject to} \; A x = z_0.
\end{equation}
and let $\mc{D}(f, x_0) = \bigcup_{t > 0} \{v \in \R^d: f(x_0 + t v) \leq f(x_0) \}$ denote the descent
cone of $f$ at $x_0$. Then, for any
$\eps > 0$, if $n \leq (1 - \eps) \delta(\mc{D}(f, x_0))$, with probability at least $1 - 32 \exp(-\eps^2 \delta_m)$,
$x_0$ fails to be the unique solution of \eqref{eq:randomconvexprog}.
\end{theoremApp}
\begin{bew} (Proposition \ref{prop:Amelunxen}). Denote by $\text{svec}: \sym^m \rightarrow \R^{\delta_m}$ the symmetric
vectorization map (cf.~\eqref{eq:svecSigma}), which is an isometry with respect to the Euclidean inner product on $\sym^m$ and $\R^{\delta_m}$,
and by $\text{svec}^{-1}: \R^{\delta_m} \rightarrow \sym^m$ its inverse. We can then apply Theorem \ref{theo:amelunxenmain}
to the setting of Proposition \ref{prop:Amelunxen} by using
\begin{equation*}
d = \delta_m, \quad x = \text{svec}(\Sigma), \quad x_0 = 0,  \quad f(x) = \iota_{\psd^m}(\text{svec}^{-1}(x)), \quad A = \begin{bmatrix}
\text{svec}(X_1) \\
\vdots \\
\text{svec}(X_n)
\end{bmatrix},
\end{equation*}
where $\iota_{\psd^m}$ is the convex indicator function of $\psd^m$ which takes the value $0$ if its argument
is contained in $\psd^m$ and $+\infty$ otherwise. Observe that  $\mc{D}(f, 0) = \psd^m$. It is shown in
\cite{Amelunxen2014}, Proposition 3.2, that the statistical dimension $\delta(\psd^m) = \delta_m/2$. This concludes
the proof.
\end{bew}

\section{Proof of Proposition \ref{prop:measurements_deterministic}}

Proposition \ref{prop:measurements_deterministic} follows from the dual problem of
the convex optimization problem associated with $\tau^2(\mc{X}, R)$. Below, it will be shown that
the Lagrangian dual of the optimization problem
\begin{align}\label{eq:primal}
\begin{split}
&\min_{A, B} \, \frac{1}{n^{1/2}} \nnorm{\mc{X}(A) - \mc{X}(B)}_2 \\
& \text{subject to} \; \; A \gec 0, \; B \gec 0, \; \tr(A) = R, \; \, \tr(B) = 1.
\end{split}
\end{align}
is given by
\begin{align}\label{eq:dual}
\begin{split}
&\max_{\theta, \delta, a} \theta \cdot R - \delta \\
&\text{subject to} \; \; \frac{\mc{X}^*(a)}{\sqrt{n}} \gec \theta I, \qquad
\frac{\mc{X}^*(a)}{\sqrt{n}} \lec \delta I, \quad \nnorm{a}_2 \leq 1.\\
\end{split}
\end{align}
The assertion of Proposition \ref{prop:measurements_deterministic} follows immediately from
\eqref{eq:dual} by identifying $\theta = \lambda_{\min}(n^{-1/2} \mc{X}^*(a))$ and $\delta = \lambda_{\max}(n^{-1/2} \mc{X}^*(a))$.
In the remainder of the proof, duality of \eqref{eq:primal} and \eqref{eq:dual} is established. Using the shortcut $\wt{\mc{X}} = \mc{X}/\sqrt{n}$, the Lagrangian of the dual problem \eqref{eq:dual} is given by
\begin{equation*}
L(\theta, \delta, a; A, B, \kappa) = \theta \cdot R - \delta  + \scp{\wt{\mc{X}}^*(a) - \theta I}{A} - \scp{\wt{\mc{X}}^*(a) -
  \delta I}{B} - \kappa (\nnorm{a}_2^2 - 1).
\end{equation*}
Taking derivatives w.r.t.~$\theta, \delta, r$ and the setting the result equal to zero, we obtain from the KKT conditions that a primal-dual optimal pair
$(\wh{\theta}, \wh{\delta}, \wh{a}, \wh{A}, \wh{B}, \wh{\kappa})$ obeys
\begin{align}\label{eq:stationarity}
\tr(\wh{A}) = R, \qquad \tr(\wh{B}) = 1, \qquad \wt{\mc{X}}(\wh{A}) - \wt{\mc{X}}(\wh{B}) - \wh{\kappa} 2 \wh{a} = 0.
\end{align}
Taking the inner product of the rightmost equation with $\wh{a}$, we obtain
\begin{align*}
&\quad \scp{\wh{a}}{\wt{\mc{X}}(\wh{A}) - \wt{\mc{X}}(\wh{B})} - \wh{\kappa} 2
\nnorm{\wh{a}}_2^2 = 0. \\
\Leftrightarrow &\quad \scp{\wt{\mc{X}}^*(\wh{a})}{\wh{A} - \wh{B}} - \wh{\kappa} 2
\nnorm{\wh{a}}_2^2 = 0. \\
\Leftrightarrow &\quad \wh{\theta} \tr(\wh{A}) - \wh{\delta} \tr(\wh{B}) - \wh{\kappa} 2
\nnorm{\wh{a}}_2^2 = 0. \\
\Leftrightarrow &\quad \wh{\theta} R - \wh{\delta} = \wh{\kappa} 2
\nnorm{\wh{a}}_2^2,
\end{align*}
where the second equivalence is by complementary slackness. Consider first
the case $\wh{\theta} R - \wh{\delta} > 0$. This entails $\wh{\kappa} > 0$ and
thus $\nnorm{\wh{a}}_2^2 = 1$, so that $2 \wh{\kappa} = \wh{\theta} R -
\wh{\delta}$. Substituting this result into the rightmost equation in \eqref{eq:stationarity} and taking norms, we obtain
\begin{equation}\label{eq:duality}
\wh{\theta} R -
\wh{\delta} = \nnorm{\wt{\mc{X}}(\wh{A}) - \wt{\mc{X}}(\wh{B})}_2 =
\frac{1}{\sqrt{n}} \nnorm{\mc{X}(\wh{A}) - \mc{X}(\wh{B})}_2.
\end{equation}
For the second case, note that $\wh{\theta} R - \wh{\delta}$ cannot be negative as $a = 0$ is
feasible for \eqref{eq:dual}. Thus, $\wh{\theta} R - \wh{\delta} = 0$ implies that $\wh{a} = 0$ and
in turn also \eqref{eq:duality}.

\section{Proof of Corollary \ref{corro:randomcov}}

The corollary follows from Proposition \ref{prop:measurements_deterministic} by choosing $a = 1/\sqrt{n}$
so that $n^{-1/2} \mc{X}^*(a) = \frac{1}{n} \su X_i$, and using that $\nnorm{\Gamma - \wh{\Gamma}_n}_{\infty} \leq \epsilon_n$
implies that $|\lambda_j(\Gamma) - \lambda_j(\wh{\Gamma}_n)| \leq \epsilon_n$, $j=1,\ldots,m$ (\cite{HornJohnson}, \S 4.3).
The specific values of $R_*$ and $\tau_*^2$ are obtained by choosing $\zeta = 2$ in Proposition \ref{prop:measurements_deterministic}.

\section{Proof of Theorem \ref{theo:slowratebound}}\label{app:slowratebound}

The following lemma is a crucial ingredient in the proof. In the sequel, let $\wh{\Delta} = \wh{\Sigma} - \Sigma^*$.
Let the eigendecomposition of $\wh{\Delta}$ be given by
\begin{equation}\label{eq:pos_neg_parts}
\wh{\Delta} = \sum_{j = 1}^m \lambda_j(\wh{\Delta}) u_j u_j^{\T} =
\underbrace{\sum_{j = 1}^m \max\{0,\lambda_j(\wh{\Delta}) \} u_j u_j^{\T}}_{\invcoloneq \wh{\Delta}^+} +
\underbrace{\sum_{j = 1}^m \min\{0,\lambda_j(\wh{\Delta}) \} u_j u_j^{\T}}_{\invcoloneq \wh{\Delta}^-}
= \wh{\Delta}^+ + \wh{\Delta}^- \\
\end{equation}
\begin{lemmaApp}\label{lem:ell1bound} Consider the decomposition \eqref{eq:pos_neg_parts}. We have $\nnorm{\wh{\Delta}^-}_1 \leq \nnorm{\Sigma^*}_1$.
\end{lemmaApp}
\begin{bew} Write $\wh{\Delta}^+ = U_+ \Lambda_+ U_+^{\T}$ and $\wh{\Delta}^- = U_- \Lambda_- U_-^{\T}$ for the eigendecompositions of $\wh{\Delta}^+$ and $\wh{\Delta}^-$, respectively. Since $\wh{\Sigma} \gec 0$,
we must have $\tr(\wh{\Sigma} U_- U_-^{\T}) \geq 0$
  and thus
\begin{align*}
0 \leq \tr(\wh{\Sigma} U_- U_-^{\T}) &= \tr(U_-^{\T} \wh{\Sigma} U_-) \\
                              &= \tr(U_-^{\T} (\Sigma^* + \wh{\Delta}) U_-)  \\
                              &= \tr(U_-^{\T} (\Sigma^* + U_+ \Lambda_+ U_+^{\T} +
                              U_- \Lambda_- U_-^{\T}) U_-) \\
                              &= \tr(\Sigma^* U_- U_-^{\T}) + \tr(\Lambda_-),
\end{align*}
where for the last identity, we have used that $U_+^{\T} U_{-} = 0$.
It follows that
\begin{equation*}
\nnorm{\wh{\Delta}^-}_1 = \norm{\Lambda_-}_1 = -\tr(\Lambda_-) \leq \tr(\Sigma^* U_-
U_-^{\T}) \leq \nnorm{\Sigma^*}_1 \nnorm{U_- U_-^{\T}}_{\infty} = \nnorm{\Sigma^*}_1.
\end{equation*}
\end{bew}
Equipped with Lemma \ref{lem:ell1bound}, we turn to the proof of Theorem
\ref{theo:slowratebound}.
\begin{bew}(Theorem \ref{theo:slowratebound}) By definition of $\wh{\Sigma}$, we have $\nnorm{y - \mc{X}(\wh{\Sigma})}_2^2 \leq \nnorm{y - \mc{X}(\Sigma^*)}_2^2$. Using \eqref{eq:tracereg_spd} and the definition of $\wh{\Delta}$, we obtain
after re-arranging terms that
\begin{align}
&\frac{1}{n} \nnorm{\mc{X}(\wh{\Delta})}_2^2 \leq \frac{2}{n} \scp{\eps}{\mc{X}(\wh{\Delta})} = \frac{2}{n} \scp{\mc{X}^*(\eps)}{\wh{\Delta}} \notag \\
\Rightarrow \quad&\frac{1}{n} \nnorm{\mc{X}(\wh{\Delta})}_2^2 \leq 2 \nnorm{\mc{X}^*(\eps)/n}_{\infty} \nnorm{\wh{\Delta}}_1 = 2 \lambda_0 (\nnorm{\wh{\Delta}^+}_1 + \nnorm{\wh{\Delta}^-}_1), \label{eq:basicinequality}
\end{align}
where we have used H\"older's inequality, the decomposition of $\wh{\Delta}$ as in Lemma \ref{lem:ell1bound} and $\lambda_0 = \nnorm{\mc{X}^*(\eps)/n}_{\infty}$. We now upper bound the l.h.s.~of \eqref{eq:basicinequality} by invoking Condition \ref{cond:slowrate} and Lemma \ref{lem:ell1bound}, which yields $\nnorm{\wh{\Delta}^-}_1 \leq \nnorm{\Sigma^*}_1$. If
$\nnorm{\wh{\Delta}^+}_1 \leq R_* \nnorm{\wh{\Delta}^-}_1$, we have
\begin{equation*}
  \frac{1}{n} \nnorm{\mc{X}(\wh{\Sigma}) - \mc{X}(\Sigma^*)}_2^2 =
\frac{1}{n} \nnorm{\mc{X}(\wh{\Delta})}_2^2 \leq 2 (R_* + 1) \lambda_0 \nnorm{\Sigma^*}_1,
\end{equation*}
which is the first part in the maximum of the bound to be established. In the opposite case,
suppose first that $\nnorm{\wh{\Delta}^-}_1 > 0$ (the case $\nnorm{\wh{\Delta}^-}_1 = 0$ is discussed
at the end of this proof) and we have $\nnorm{\wh{\Delta}^+}_1 / \nnorm{\wh{\Delta}^-}_1 = \wh{R} > R_*> 1$.
Consequently,
\begin{align*}
\frac{1}{n} \nnorm{\mc{X}(\wh{\Delta})}_2^2 &= \frac{1}{n} \nnorm{\mc{X}(\wh{\Delta}^+)
  - \mc{X}(-\wh{\Delta}^-)}_2^2 \\
&= \nnorm{\wh{\Delta}^-}_1^2 \; \frac{1}{n} \norm{\mc{X}\left(\frac{\wh{\Delta}^+}{\nnorm{\wh{\Delta}^-}_1} \right)
  - \mc{X} \left(\frac{-\wh{\Delta}^-}{\nnorm{\wh{\Delta}^-}_1} \right)}_2^2 \\
&\geq  \nnorm{\wh{\Delta}^-}_1^2 \, \min_{\substack{A \in \wh{R} \mc{S}_1^+(m) \\
B \in \mc{S}_1^+(m)}} \, \frac{1}{n} \nnorm{\mc{X}(A) - \mc{X}(B)}_2^2 \\
&=\tau^2(\mc{X}, \wh{R})  \nnorm{\wh{\Delta}^-}_1^2 = \tau^2(\mc{X}, \wh{R}) \frac{\nnorm{\wh{\Delta}^+}_1^2}{\wh{R}^2} \\
\end{align*}
Inserting this into \eqref{eq:basicinequality}, we obtain the following upper bound on
$\nnorm{\wh{\Delta}^+}_1$.
\begin{align*}
&\frac{\tau^2(\mc{X}, \wh{R})}{\wh{R}^2} \nnorm{\Delta^+}_1^2 \leq 2
\lambda_0 \frac{\wh{R} + 1}{\wh{R}} \nnorm{\wh{\Delta}^+}_1 \\
\Rightarrow & \quad \nnorm{\wh{\Delta}^+}_1 \leq 2
\lambda_0 \frac{\wh{R} (\wh{R}+1)}{\tau^2(\mc{X}, \wh{R})} \leq 4 \lambda_0 \frac{\wh{R}^2}{\tau^2(\mc{X}, \wh{R})} \leq 4 \lambda_0 \frac{R_*^2}{\tau_*^2},
\end{align*}
where the last inequality follows from the observation that for any $R \geq R_*$
\begin{equation*}
\tau^2(\mc{X}, R) \geq (R/R_*)^2 \tau^2(\mc{X}, R_*),
\end{equation*}
which can be easily seen from the dual problem \eqref{eq:dual} associated with $\tau^2(\mc{X}, R)$.
Substituting the above bound on $\nnorm{\wh{\Delta}^+}_1$ into \eqref{eq:basicinequality} and using the bound $\nnorm{\wh{\Delta}^-}_1 \leq \nnorm{\Sigma^*}_1$ yields the second part in the maximum of the desired bound. To finish the proof,
we still need to address the case $\nnorm{\wh{\Delta}^-}_1 = 0$. Recalling the definition
of the quantity $\tau_0^2(\mc{X})$ in \eqref{eq:selfreg}, we bound
\begin{align*}
\frac{1}{n} \nnorm{\wh{X}(\wh{\Delta})}_2^2 = \frac{1}{n} \nnorm{\wh{X}(\wh{\Delta}^+)}_2^2 \geq \tau_0^2(\mc{X}) \nnorm{\wh{\Delta}^+}_1^2.
\end{align*}
Inserting this into \eqref{eq:basicinequality}, we obtain from \eqref{eq:selfreg_weaker}
\begin{equation*}
\nnorm{\wh{\Delta}^+}_1 \leq  \frac{2\lambda_0}{\tau_0^2(\mc{X})} \leq \frac{2 \lambda_0 (R_{*} - 1)^2}{\tau_*^2}.
\end{equation*}
Back-substitution into \eqref{eq:basicinequality} yields a bound that is implied by that of Theorem \ref{theo:slowratebound}.
This concludes the proof.
\end{bew}
\emph{Bound on $\lambda_0$.} The bound on $\lambda_0$ is an application of Theorem 4.6.1 in \cite{Tropp2014}.
\begin{theoremApp} \cite{Tropp2014} Consider a sequence $\{ X_i \}_{i = 1}^n$ of fixed
matrices in $\sym^m$ and let $\{ \eps_i \}_{i = 1}^n  \overset{\text{i.i.d.}}{\sim} N(0, \sigma^2)$. Then for all $t \geq 0$
\begin{equation*}
\p \left(\norm{\sum_{i = 1}^n \eps_i X_i}_{\infty} \geq t \right) \leq 2 m
\exp(-t^2 / (2 \sigma^2 V^2)),
\quad V^2 \coloneq \norm{\su X_i^2}_{\infty}.
\end{equation*}
\end{theoremApp}
Choosing $t = \sigma V \sqrt{(1 + \mu) 2 \log(2m)}$ yields the desired bound.

\section{Proof of Theorem \ref{theo:slowratebound}, Remark 3}\label{sec:Vn}

The bound hinges on the following concentration result for the extreme eigenvalues of
the sample covariance of a Gaussian sample.
\begin{theoremApp}\label{theo:concentration_spectrum}\cite{DavidsonSzarek} Let $z_1,\ldots,z_N$ be an i.i.d.~sample from $N(0, I_m)$ and
let $\Gamma_N = \frac{1}{N} \sum_{i = 1}^N z_i z_i^{\T}$. We then have for any $\delta > 0$
\begin{align*}
&\p \left(\lambda_{\max} \left(\frac{1}{N} \Gamma_N  \right) >  \left( 1 + \delta + \sqrt{\frac{m}{N}} \right)^2 \right) \leq \exp(- N\delta^2 / 2).
\end{align*}
\end{theoremApp}
In the proof, we also make use of the following fact.
\begin{lemmaApp}
Let $\{ X_i \}_{i = 1}^n \subset \psd^m$. Then
\begin{equation*}
\norm{\su X_i^2}_{\infty} \leq \max_{1 \leq i \leq n} \norm{X_i}_{\infty} \norm{\su X_i}_{\infty}.
\end{equation*}
\end{lemmaApp}
\begin{bew} First note that for any $v \in \R^m$ and any $M \in \psd^m$, we have that
\begin{equation*}
v^{\T} M^2 v  = \sum_{j = 1}^m \lambda_j^2(M) (u_j^{\T} v)^2 \leq
\lambda_{\max}(M) \sum_{j=1}^m \lambda_j(M) (u_j^{\T} v)^2 = \nnorm{M}_{\infty}  v^{\T} X v,
\end{equation*}
where $\{u_j \}_{j = 1}^m$ are the eigenvectors of $X$. Accordingly, we have
\begin{align*}
\norm{\su X_i^2}_{\infty} = \max_{\nnorm{v}_2 = 1} v^{\T} \sum_{i = 1}^n X_i^2 v
&\leq \max_{1 \leq i \leq n} \nnorm{X_i}_{\infty}  \max_{\nnorm{v}_2 = 1} v^{\T} \su X_i v \\
&= \max_{1 \leq i \leq n} \nnorm{X_i}_{\infty} \norm{\su X_i}_{\infty}.
\end{align*}
\end{bew}
We now establish the bound to be shown. Each measurement matrix can be expanded as
\begin{equation*}
X_i = \frac{1}{q} \sum_{k = 1}^q z_{ik} z_{ik}^{\T}, \quad \{z_{ik} \}_{k = 1}^q \overset{\text{i.i.d.}}{\sim} N(0,I_m), \; \, i=1,\ldots,n.
\end{equation*}
Accordingly, we have
\begin{align*}
\norm{\frac{1}{n} \su X_i^2}_{\infty} &=\norm{\frac{1}{n} \su \left\{ \frac{1}{q} \sum_{k = 1}^q z_{ik} z_{ik}^{\T} \right\}^2}_{\infty} \\
                                    &\leq \max_{1 \leq i \leq n} \left\{\norm{\left\{ \frac{1}{q} \sum_{k = 1}^q z_{ik} z_{ik}^{\T} \right\}}_{\infty} \right\}
                                     \norm{\frac{1}{nq} \sum_{i = 1}^n \sum_{k = 1}^q z_{ik} z_{ik}^{\T}}_{\infty} \\
&\leq   \max_{1 \leq i \leq n} \left\{\lambda_{\max} \left( \frac{1}{q} \sum_{k = 1}^q z_{ik} z_{ik}^{\T} \right) \right\} \lambda_{\max}(\Gamma_{nq})
\end{align*}
where $\Gamma_{nq}$ follows the distribution of $\Gamma_N$ in Theorem \ref{theo:concentration_spectrum}  with $N = nq$. For the
first term, applying Theorem \ref{theo:concentration_spectrum} with $N = q$ and $\delta = \sqrt{4 m \log(n) / q}$ and using the
union bound, we obtain that
\begin{equation*}
\p \left(\lambda_{\max} \left( \frac{1}{q} \sum_{k = 1}^q z_{ik} z_{ik}^{\T} \right) > \left( \frac{\sqrt{q} + \sqrt{m} + \sqrt{4m \log n}}{\sqrt{q}} \right)^2 \right) \leq \exp(-(2m - 1) \log n).
\end{equation*}
Applying Theorem \ref{theo:concentration_spectrum} to $\Gamma_N$ with $\delta = 1/\sqrt{q}$, we obtain that
\begin{equation*}
\p \left(\lambda_{\max}(\Gamma_{nq}) > \left(1 + \frac{1}{\sqrt{q}} + \sqrt{\frac{m}{nq}} \right)^2 \right) \leq \exp(-n/2).
\end{equation*}
Combining the two previous bounds yields the assertion.

\section{Proof of Proposition \ref{prop:recovery_noiseless}}

In the sequel, we write $\Pi_{\TT}$ and $\Pi_{\TT^{\perp}}$ for the  orthogonal projections on $\TT$ and $\TT^{\pe}$, respectively.
Note first that since the $\{ \eps_i \}_{i = 1}^n$ are zero, any minimizer $\wh{\Sigma}$ satisfies
\begin{align}\label{eq:noiseless_fit}
\mc{X}(\wh{\Sigma}) = \mc{X}(\Sigma^*) \; \Longleftrightarrow \mc{X}(\wh{\Delta}) = 0  \Longleftrightarrow
\mc{X}(\wh{\Delta}_{\TT}) + \mc{X}(\wh{\Delta}_{\TT^{\pe}}) = 0
\end{align}
where $\wh{\Delta}_{\TT} = \Pi_{\TT} \wh{\Delta}$ and $\wh{\Delta}_{\TT^{\pe}} = \Pi_{\TT^{\pe}} \wh{\Delta}$, where we recall
that $\wh{\Delta} = \wh{\Sigma} - \Sigma^*$. Note that
since $\Sigma^* = \Pi_{\TT} \Sigma^*$, for $\wh{\Sigma}$ to be feasible, it is necessary that
$\wh{\Delta}_{\TT^{\pe}} \gec 0$.

Suppose first that $\tau^2(\TT) = 0$. Then there exist $\Theta \in \TT$ and $\Lambda \in \mc{S}_1^+(m) \cap
\TT^{\pe}$ such that $\mc{X}(\Theta) + \mc{X}(\Lambda) = 0$. Hence, for any $\Sigma^* \in \TT$ with
$\Sigma^* + \Theta \gec 0$, the choices $\wh{\Delta}_{\TT} = \Theta$ and $\wh{\Delta}_{\TT^{\pe}} = \Lambda$ ensure
that $\wh{\Sigma}$ is feasible and that \eqref{eq:noiseless_fit} is satisfied. Since $\Lambda$ is contained in the Schatten 1-norm
sphere of radius $1$, it is necessarily non-zero and thus $\wh{\Sigma} \neq \Sigma^*$.\\
If $\phi^2(\TT) = 0$, there exists $0 \neq \Theta \in \TT$ such that $\mc{X}(\Theta) = 0$. Consequently, for any
$\Sigma^* \in \TT \cap \psd^m$ with $\wh{\Sigma} = \Sigma^* + \Theta \gec 0$, \eqref{eq:noiseless_fit} is satisfied with
$\wh{\Sigma} \neq \Sigma^*$.

Conversely, if $\tau^2(\TT) > 0$, \eqref{eq:noiseless_fit} cannot be satisfied for $\wh{\Delta}_{\TT^{\pe}} \gec 0$,
$\wh{\Delta}_{\TT^{\pe}} \neq 0$. Otherwise, we could divide by $\tr(\wh{\Delta}_{\TT^{\pe}})$, which would yield
\begin{equation*}
\mc{X}(\underbrace{\wh{\Delta}_{\TT} \big / \tr(\wh{\Delta}_{\TT^{\pe}})}_{\in \TT}) +
\mc{X}(\underbrace{\wh{\Delta}_{\TT^{\pe}} \big / \tr(\wh{\Delta}_{\TT^{\pe}})}_{\in \mc{S}_1^+(m) \cap \TT^{\pe}}) = 0,
\end{equation*}
which would imply $\tau^2(\TT) = 0$. Therefore, we must have $\wh{\Delta}_{\TT^{\pe}} = 0$
and $\mc{X}(\wh{\Delta}_{\TT})  = 0$, which implies $\wh{\Delta}_{\TT} = 0$ as long as $\phi^2(\TT) > 0$.

\section{Proof of Theorem \ref{theo:estimationerror}}

Let $\wh{\Delta} = \wh{\Sigma} - \Sigma^*$,
$\wh{\Delta}_{\TT} = \Pi_{\TT} \wh{\Delta}$ and $\wh{\Delta}_{\TT^{\pe}} = \Pi_{\TT^{\pe}} \wh{\Delta} \gec 0$ as in the preceding
proof. We start with the following analog to \eqref{eq:basicinequality}
\begin{align}\label{eq:basicinequality_TT}
\frac{1}{n} \nnorm{\mc{X}(\wh{\Delta})}_2^2 = \frac{1}{n} \nnorm{\mc{X}(\wh{\Delta}_{\TT} + \wh{\Delta}_{\TT^{\pe}})}_2^2 \leq 2 \lambda_0 (\nnorm{\wh{\Delta}_{\TT}}_1 + \nnorm{\wh{\Delta}_{\TT^{\pe}}}_1)
\end{align}
Suppose that $\wh{\Delta}_{\TT^{\pe}} \neq 0$. We then have
\begin{align*}
\nnorm{\wh{\Delta}_{\TT^{\pe}}}_1^2  \; \left\{
\frac{1}{n} \norm{\mc{X} \left( \frac{\wh{\Delta}_{\TT}}{\nnorm{\wh{\Delta}_{\TT^{\pe}}}_1}\right) +
\mc{X} \left(\frac{\wh{\Delta}_{\TT^{\pe}}}{\nnorm{\wh{\Delta}_{\TT^{\pe}}}_1} \right)}_2^2  \right\}
  \leq 2 \lambda_0 (\nnorm{\wh{\Delta}_{\TT}}_1 + \nnorm{\wh{\Delta}_{\TT^{\pe}}}_1)
\end{align*}
Since $\wh{\Delta}_{\TT} / \nnorm{\wh{\Delta}_{\TT^{\pe}}}_1 \in \TT$ and
$\wh{\Delta}_{\TT^{\pe}} / \nnorm{\wh{\Delta}_{\TT^{\pe}}}_1 = \wh{\Delta}_{\TT^{\pe}} / \tr(\wh{\Delta}_{\TT^{\pe}}) \in \mc{S}_1^+(m)$,
we obtain that the term inside the curly brackets is lower bounded by $\tau^2(\mathbb{T})$ and thus
\begin{equation}\label{eq:Deltaperp_bound}
\nnorm{\wh{\Delta}_{\TT^{\pe}}}_1 \leq \frac{2 \lambda_0}{\tau^2(\TT)} \left(1 + \frac{\nnorm{\wh{\Delta}_{\TT}}_1}{\nnorm{\wh{\Delta}_{\TT^{\pe}}}_1} \right)
\end{equation}
On the other hand, expanding the quadratic term in \eqref{eq:basicinequality_TT}, we obtain that
\begin{align}
& \frac{1}{n} \nnorm{\mc{X}(\wh{\Delta}_{\TT})}_2^2 - \frac{2}{n} \scp{\mc{X}(\wh{\Delta}_{\TT})}{\mc{X}(\wh{\Delta}_{\TT^{\pe}})}  \leq \frac{1}{n} \nnorm{\mc{X}(\wh{\Delta})}_2^2 \leq 2 \lambda_0 (\nnorm{\wh{\Delta}_{\TT}}_1 + \nnorm{\wh{\Delta}_{\TT^{\pe}}}_1) \notag \\
\Rightarrow \quad& \frac{1}{n} \nnorm{\mc{X}(\wh{\Delta}_{\TT})}_2^2 \leq 2 \lambda_0 (\nnorm{\wh{\Delta}_{\TT}}_1  + \nnorm{\wh{\Delta}_{\TT^{\pe}}}_1) + 2 \mu(\TT)  \nnorm{\wh{\Delta}_{\TT}}_1 \nnorm{\wh{\Delta}_{\TT^{\pe}}}_1 \notag \\
\Rightarrow \quad&\phi^2(\TT)  \nnorm{\wh{\Delta}_{\TT}}_1^2 \leq 2 \lambda_0 (\nnorm{\wh{\Delta}_{\TT}}_1  + \nnorm{\wh{\Delta}_{\TT^{\pe}}}_1) + 2 \mu(\TT)  \nnorm{\wh{\Delta}_{\TT}}_1 \nnorm{\wh{\Delta}_{\TT^{\pe}}}_1 \notag \\
\Rightarrow \quad& \nnorm{\wh{\Delta}_{\TT}}_1 \leq \frac{2 \lambda_0 \left( 1 + \nnorm{\wh{\Delta}_{\TT^{\pe}}}_1 \big / \nnorm{\wh{\Delta}_{\TT}}_1 \right) + 2 \mu(\TT) \nnorm{\wh{\Delta}_{\TT^{\pe}}}_1}{\phi^2(\TT)} \label{eq:DeltaT_bound}
\end{align}
We now distinguish several cases.

\textbf{Case 1}: $\nnorm{\wh{\Delta}_{\TT}}_1 \leq \nnorm{\wh{\Delta}_{\TT^{\pe}}}_1$. It then immediately follows from
\eqref{eq:Deltaperp_bound} that
\begin{equation}\label{eq:ell1bound_1}
\nnorm{\wh{\Delta}}_1 \leq \frac{8 \lambda_0}{\tau^2(\TT)} \invcoloneq T_3.
\end{equation}
\textbf{Case 2a}: $\nnorm{\wh{\Delta}_{\TT}}_1 > \nnorm{\wh{\Delta}_{\TT^{\pe}}}_1$ and $\nnorm{\wh{\Delta}_{\TT^{\pe}}}_1 \leq 4 \lambda_0 / \phi^2(\TT)$. From \eqref{eq:DeltaT_bound}, we first get
\begin{equation} \label{eq:DeltaT_bound_plugin}
 \nnorm{\wh{\Delta}_{\TT}}_1 \leq
\frac{4 \lambda_0  + 2 \mu(\TT) \nnorm{\wh{\Delta}_{\TT^{\pe}}}_1}{\phi^2(\TT)}
\end{equation}
and thus
\begin{equation}\label{eq:ell1bound_2}
\nnorm{\wh{\Delta}}_1 \leq \frac{8 \lambda_0}{\phi^2(\TT)} \left( 1 + \frac{\mu(\TT)}{\phi^2(\TT)}  \right ) \invcoloneq T_2
\end{equation}
\textbf{Case 2b}:  $\nnorm{\wh{\Delta}_{\TT}}_1 > \nnorm{\wh{\Delta}_{\TT^{\pe}}}_1$ and $\nnorm{\wh{\Delta}_{\TT^{\pe}}}_1 > 4 \lambda_0 / \phi^2(\TT)$. Plugging \eqref{eq:DeltaT_bound_plugin} into \eqref{eq:Deltaperp_bound}, we obtain that
\begin{equation*}
\nnorm{\wh{\Delta}_{\TT^{\pe}}}_1 \leq \frac{4 \lambda_0}{\tau^2(\TT)} + \frac{4 \lambda_0 \mu(\TT)}{\tau^2(\TT) \phi^2(\TT)}.
\end{equation*}
Substituting this bound back into \eqref{eq:DeltaT_bound_plugin} yields
\begin{equation*}
\nnorm{\wh{\Delta}_{\TT}}_1 \leq \frac{4 \lambda_0}{\phi^2(\TT)} + \frac{8 \lambda_0 \mu(\TT)}{\tau^2(\TT) \phi^2(\TT)}  +
\frac{8 \lambda_0 \mu^2(\TT)}{\phi^4(\TT) \tau^2(\TT)}.
\end{equation*}
Collecting terms, we obtain altogether
\begin{equation}\label{eq:ell1bound_3}
\nnorm{\wh{\Delta}}_1 \leq 8 \lambda_0 \frac{\mu(\TT)}{\tau^2(\TT) \phi^2(\TT)} \left(\frac{3}{2} + \frac{\mu(\TT)}{\phi^2(\TT)}   \right) + 4 \lambda_0 \left(\frac{1}{\phi^2(\TT)} + \frac{1}{\tau^2(\TT)} \right) \invcoloneq T_1.
\end{equation}
Combining \eqref{eq:ell1bound_1}, \eqref{eq:ell1bound_2} and \eqref{eq:ell1bound_3} yields the assertion.




\end{document}